%% file: ref.tex
\documentclass[journal]{IEEEtran}

\input{preamble}

\begin{document}

\title{PointDifformer: Robust Point Cloud Registration with Neural Diffusion and Transformer}

\author{
        Rui~She$^{*}$,
        Qiyu~Kang$^{*}$,
        Sijie~Wang$^{*}$,
        Wee~Peng~Tay,~\IEEEmembership{Senior~Member,~IEEE,} 
        Kai~Zhao, 
        Yang~Song, \\
        Tianyu~Geng, 
        Yi~Xu,
        Diego~Navarro~Navarro and 
        Andreas~Hartmannsgruber
\thanks{$^{*}$R. She, Q. Kang and S. Wang, contributed equally to this work.} 
\thanks{R. She, Q. Kang, S. Wang, W. P. Tay, K. Zhao, Y. Song, T. Geng, and Y. Xu are with Nanyang Technological University, Singapore. {\{rui.she@; qiyu.kang@; wang1679@e.; kai.zhao@; songy@; tianyu.geng@; yi.xu@; wptay@\}ntu.edu.sg}
        }
\thanks{
        D. N. Navarro and A. Hartmannsgruber are with Continental Automotive Singapore Pte. Ltd., Singapore. {\{diego.navarro.navarro; andreas.hartmannsgruber\}@continental.com}
        }
}




\maketitle

\begin{abstract}
Point cloud registration is a fundamental technique in 3D computer vision with applications in graphics, autonomous driving, and robotics. However, registration tasks under challenging conditions, under which noise or perturbations are prevalent, can be difficult. We propose a robust point cloud registration approach that leverages graph neural Partial Differential Equations (PDEs) and heat kernel signatures.
Our method first uses graph neural PDE modules to extract high-dimensional features from point clouds by aggregating information from the 3D point neighborhood, thereby enhancing the robustness of the feature representations. Then, we incorporate heat kernel signatures into an attention mechanism to efficiently obtain corresponding keypoints. Finally, a Singular Value Decomposition module with learnable weights is used to predict the transformation between two point clouds.
Empirical experiments on a 3D point cloud dataset demonstrate that our approach not only does achieve state-of-the-art performance for point cloud registration but also exhibits better robustness to additive noise or 3D shape perturbations. 
\end{abstract}


\section{Introduction}

\IEEEPARstart{I}{n} the era of intelligent and smart perception, 3D computer vision techniques are increasingly being used in various fields, such as autonomous driving, robotics, and scene modeling \cite{zhang2022efficient,quan2020compatibility,Kang22itsc}. Point cloud registration is a crucial task in 3D computer vision and has become an important tool in many applications, including object detection, odometry estimation, and SLAM \cite{sun2022weakly, zhou2021t, shi2020point, shan2020lio, zhang2014loam}, owing to its robustness against seasonal changes and illumination variations. Point cloud registration aims to estimate the transformation or relative pose between two given 3D point cloud frames \cite{wang2019deep}.

Iterative algorithms are widely used for point cloud registration \cite{besl1992method, yang2013go, koide2021voxelized}. The Iterative Closest Point (ICP) algorithm is a well-known iterative method for point cloud registration that matches the closest points between two point clouds, iteratively updating the transformation matrix until convergence \cite{besl1992method}. ICP has been successfully used in numerous fields, including robotic perception and autonomous driving.

Despite their usefulness, iterative algorithms face challenges that limit their effectiveness in certain scenarios. The non-convexity of the optimization problem presents a significant challenge, making it difficult to obtain the global optimum \cite{wang2019deep}. As a result, iterative algorithms may converge to sub-optimal solutions, especially in complex and non-rigid scenes. Additionally, the performance of iterative algorithms heavily relies on the initialization of the algorithm, which can be time-consuming and computationally expensive.
Sparse and non-uniform point clouds present another significant challenge for iterative algorithms in finding corresponding point pairs between two point clouds. Traditional approaches, such as nearest-neighbor search, may fail to find matching pairs in such cases, leading to errors in the registration result \cite{wang2019deep, wei2020end}.

To address these challenges, deep learning-based methods \cite{wang2019deep, wei2020end, xu2022glorn, zhang2022ps} have been developed for predicting transformation matrices or relative poses. These methods are designed for various scenarios, including indoor and outdoor environments \cite{lu2021hregnet, choy2019fully, bai2020d3feat, ao2021spinnet, yu2021cofinet}. However, robust point cloud registration remains a challenging problem due to factors such as LiDAR scan distortion, dynamic objects, and environmental noise \cite{li2021point, wang2020multientity, chen2019plade, yu2019advanced, YananZhao2023}. Efficiently estimating the transformation under scenarios with additive noise and perturbations remains an open problem.

In this paper, we propose a model for point cloud registration that utilizes a robust feature descriptor based on graph neural diffusion. We also present an end-to-end transformation estimation method by introducing the heat kernel signature into the attention module, without any prior prediction information. Our approach attempts to address the challenges faced by iterative algorithms, leading to robust and efficient point cloud registration.
Our proposed approach is motivated by the following:
\begin{itemize}
\item Graph neural PDE learning has demonstrated robustness for representing graph-structured data, as highlighted in \cite{song2022robustness}. Our aim is to leverage this module for effective point cloud representation by constructing a neighborhood graph in the feature space.
\item We believe that the shape isometry-invariance of the heat kernel signature, as described in \cite{sun2009concise}, makes it beneficial to incorporate into attention mechanisms for improved robustness from a shape-preserving perspective.
\end{itemize}

Our main contributions are as follows:
\begin{itemize}
\item We design a 3D point cloud representation module based on graph neural PDE learning.
\item We propose a robust 3D point cloud registration method using the graph neural diffusion modules and the attention mechanism with a heat kernel signature.
\item We empirically demonstrate that our point cloud registration method outperforms other baselines not only under normal scenarios but also when noise and perturbations are present.
\end{itemize}

The rest of this paper is organized as follows. In \cref{sect:relatedworks}, we discuss the related works. In \cref{sect:model}, we describe our proposed model in detail. In \cref{sect:exper}, we present the experimental results to evaluate our model and compare it with several baselines. Finally, we conclude the paper in \cref{sect:conc}.

\section{Related Work}\label{sect:relatedworks}
In this section, we summarize relevant literature in the areas of point cloud registration, point cloud feature representation, and neural diffusion, including works utilizing the heat kernel signature in point cloud feature descriptors. 

\subsection{Point Cloud Registration Methods}

\textbf{Iteration-based Methods.}
Iteration-based methods, such as the Iterative Closest Point (ICP) \cite{besl1992method} and the RANdom SAmple Consensus (RANSAC) \cite{fischler1981random}, are classical approaches commonly used for point cloud registration. However, due to its slow convergence rate, RANSAC requires high computing resources and has a high running time complexity. The performance of ICP heavily relies on the accuracy of the initial value estimation, making it prone to suboptimal solutions. 
To address these challenges, several refinement methods for ICP have been proposed, such as the Branch-and-Bound (BnB) method \cite{yang2013go}, convex relaxation improvement \cite{maron2016point}, and mixed-integer programming \cite{guerout2017mixed}. However, these methods may be computationally expensive and do not ensure global optimality. 
Alternatively, updated ICP methods such as Voxelized ICP \cite{koide2021voxelized} and Generalized-ICP \cite{segal2009generalized} have been developed to improve both acceleration and accuracy. 

\textbf{Correspondence-based estimators.}
Correspondence-based estimators are commonly used for point cloud registration, which involves estimating the transformation between two frames \cite{deng2018ppf,choy2019fully,deng2018ppfnet,gojcic2019perfect}. 
This approach obtains correspondences between two point clouds and then uses pose estimators such as RANSAC \cite{fischler1981random,bai2020d3feat,huang2021predator}, Singular Value Decomposition (SVD) \cite{besl1992method,wang2019deep,wei2020end,chen2022sc2} and  Maximal Cliques (MAC) \cite{zhang20233d} to predict the transformation.
There are generally two types of correspondence-based estimators. One involves repeatable keypoint detection \cite{lu2019l3,lu2019deepvcp,bai2020d3feat,huang2021predator}, followed by using learned or handcrafted keypoint descriptors for correspondence acquisition \cite{ao2021spinnet,choy2019fully,aoki2019pointnetlk} or similarity measures to obtain the correspondences \cite{quan2020compatibility,chen2022sc2}. For example, DeepVCP \cite{lu2019deepvcp} uses PointNet++ \cite{qi2017pointnet++} to extract features for the point clouds and learns keypoint correspondences based on matching probabilities among candidates. D3Feat \cite{bai2020d3feat} employs 3D fully convolutional networks to output detection scores and descriptive features for 3D points. PREDATOR \cite{huang2021predator} uses an overlap-attention block for cross-information between two point clouds and makes good use of their overlap region to achieve registration.
The other \cite{wang2019deep,yu2021cofinet} involves correspondence retrieval for all possible matching point pairs without keypoint detection. For instance, Deep Closest Point (DCP) \cite{wang2019deep} aligns features based on the interaction of the point clouds. CoFiNet \cite{yu2021cofinet} achieves hierarchical correspondences with coarse and finer scales, without keypoint detection.
In both types of estimators, point cloud descriptors play significant roles, mainly contributing to the robustness and accuracy of the entire pipeline.

\textbf{Learning-based estimators.}
In order to achieve more robust non-handcrafted estimators, learning-based methods are introduced into the transformation prediction \cite{qin2022geometric}.
Since conventional estimators like RANSAC have drawbacks in terms of convergence speed and are unstable in the presence of numerous outliers, learning-based estimators 
\cite{choy2020deep,wang2022you,zhao2022graphreg,poiesi2022learning,tang2022multi,lee2021deep,lu2021hregnet,pais20203dregnet}, such as StickyPillars \cite{fischer2021stickypillars}, PointDSC \cite{bai2021pointdsc}, EDFNet \cite{zhang2022learning}, GeoTransformer \cite{qin2022geometric}, Lepard \cite{li2022lepard}, RoITr \cite{yu2023rotation}, BUFFER \cite{ao2023buffer} and RoReg \cite{wang2023roreg}, have attracted much interest. 
Moreover, auxiliary modules or prior information can be incorporated into learning-based estimators, such as Prior-embedded Explicit Attention Learning (PEAL) \cite{yu2023peal} and VBReg \cite{jiang2023robust}. 
Some of these classification neural networks can filter out extreme outliers and some estimation neural networks are designed to output the transformation.
From the perspective of accuracy and running efficiency, they perform better than those conventional robust estimators. 
While, they need extra neural network training, which holds more time and space complexity. In contrast, our model achieves robust and accurate registration without the need for training the estimation networks to compute the final transformation in the output.

\subsection{Point Cloud Feature Representation}

To extract more efficient features for point clouds, methods using different neural networks are studied.
In general, we can classify point cloud feature representation methods into three categories as follows.  

The first category performs voxel alignment on the points and then obtains the corresponding features based on a 3D Convolutional Neural Network (CNN) \cite{zhou2018voxelnet,sindagi2019mvx,kopuklu2019resource,kumawat2019lp}. 
In this regard, the full information in the point cloud is used to learn the representation. However, it takes more computational resources to deal with a sparse and irregular point cloud when using closely spaced 3D voxels for more precise quantization.

The second category reduces a 3D point cloud to a 2D map and then exploits the classical 2D CNN to extract features \cite{su2015multi}. 
The commonly used 2D maps are the bird's-eye view map, cylindrical map, spherical map, and camera-plane map, for which computational cost is incurred during the preprocessing stage. Due to quantization errors, this approach can also introduce unexpected noise. 

The third category is to extract features from the raw point clouds directly using specific neural networks. 
PointNet \cite{qi2017pointnet} and PointNet++ \cite{qi2017pointnet++} extract local point features independently and obtain global features through max-pooling. 
To incorporate local neighborhood information, Dynamic Graph Convolutional Neural Networks (DGCNN) \cite{wang2019dynamic} uses a dynamic graph network, and LPDNet \cite{liu2019lpd} jointly exploits the geometry space and feature space.
KPConv \cite{thomas2019kpconv} uses kernel points to achieve more flexible convolutions compared with fixed grid convolutions. 
PointGLR \cite{rao2020global} considers not only local features but also global patterns in point clouds. DIP \cite{poiesi2021distinctive} and GeDi \cite{poiesi2022learning} extract local point cloud patches based on rotation-invariant compact descriptors, which can be used in different data domains. Point Cloud Transformer (PCT) \cite{guo2021pct} utilizes the Transformer to generate permutation-invariant descriptors for point clouds.
Moreover, there are also other learning-based point cloud representation methods, such as PointMLP \cite{ma2022pointmlp}, and PointNeXt \cite{qian2022pointnext}.

\begin{figure*}[!htb]
\centering
\fbox{\includegraphics[width=0.98\linewidth]{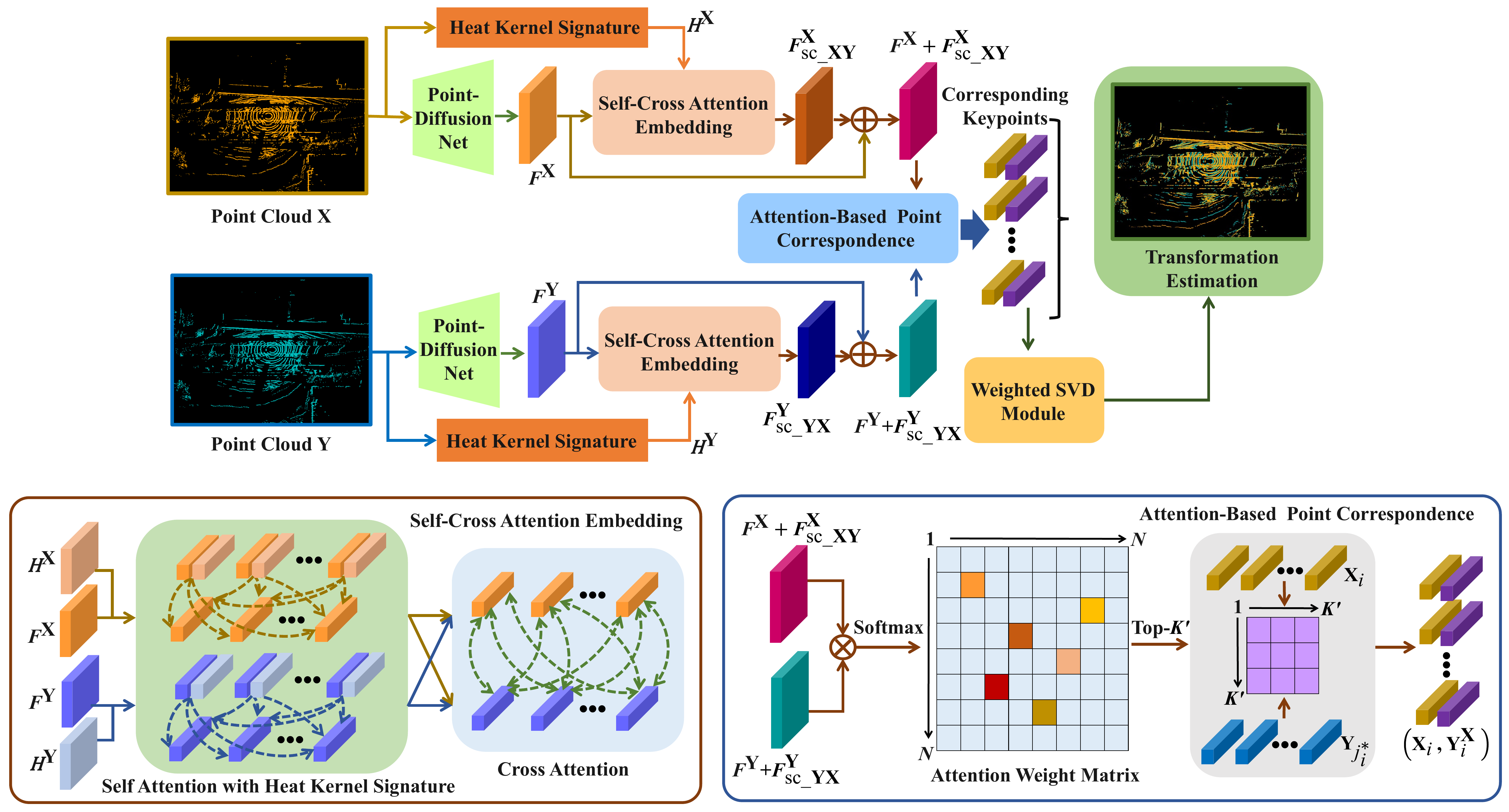}}
\caption{PointDifformer for point cloud registration. The details of the modules are provided in \cref{subsect:model_details}.}
\label{fig:model_robT}
\end{figure*}

\subsection{Neural Diffusion}
Neural diffusion methods \cite{chen2018neural,chamberlain2021grand} combine neural networks with ordinary and partial differential equations. 
For a neural Ordinary Differential Equation (ODE) layer \cite{chen2018neural} with input $\bZ(0)$ and output $\bZ(T)$, the relationship between $\bZ(0)$ and $\bZ(T)$ is given by 
\begin{align}
\ddfrac{\bZ(t)}{t}=f_{\mathrm{ODE}}(\bZ(t), t), \label{eq:NODE}
\end{align}
where $f_{\mathrm{ODE}}:\Real^n \times [0,T)\to \Real^n$ is a learnable layer and $\bZ: [0,T) \to \Real^n$ denotes the state of the neural ODE. 
At the terminal time $T \in [0,\infty)$, the output $\bZ(T)$ is given by
\begin{align}
\bZ(T) = \bZ(0) + \int_{0}^{T} f_{\mathrm{ODE}}(\bZ(t), t) {\rm d}t.
\end{align}
In this paper, we consider only the time-invariant (autonomous) case, i.e., $f_{\mathrm{ODE}}(\bZ(t), t) = f_{\mathrm{ODE}}(\bZ(t))$. 

For graph-structured data, graph neural PDEs are designed based on continuous flows, which represent the graph features more concisely and stably \cite{chen1998stability,chamberlain2021grand,chamberlain2021blend,song2022robustness,Wang2022robustloc,kang2021Neurips,zhao2020dredge,zhao2023adversarial}.
Neural ODEs/PDEs are more robust in defending perturbations and even attacks, compared with other deep neural networks without neural diffusion (cf.\ \cite{she2023robustmat,song2022robustness}). 
Compared with conventional graph neural networks (GNNs), including the Graph ATtention network (GAT) or the Graph Convolutional Network (GCN), graph neural PDEs have superior performance in some applications such as the node classification for graph-structured data.
To approximately solve the graph neural PDEs \cite{chamberlain2021grand}, the neural ODE solvers proposed in \cite{chen2018neural} can be exploited. 

\subsection{Heat Kernel Signature}
Based on the heat diffusion process, the heat kernel signature \cite{sun2009concise} is presented as an intrinsic feature and is given by
\begin{align}
    h(x, t)=\sum_{i=0}^{\infty} \exp{(-\lambda_i t)} \phi_i^2(x), \label{eq.heat_kernel}
\end{align}
where $x$ is a 3D point in a point cloud, $\lambda_i$s denote eigenvalues and $\phi_i$s are the corresponding eigenfunctions of the Laplace-Beltrami operator. 
The feature $h(x, t)$ is a robust local geometric descriptor containing large-scale information \cite{chen2019lassonet,bronstein2010scale}. 
From the physics perspective, this feature descriptor represents the temperature evolution of a point at which a heat source is placed and removed immediately. The heat diffuses to the neighborhood of the point \cite{sun2009concise,tam2012registration}.
This evolution is based on the temperature diffusion speed, which essentially depends on the geometry of the objects projected by the point clouds.    

From a geometric perspective, the heat kernel signature is isometry-invariant, meaning that two isometric shapes have equivalent heat kernel signatures. 
If the heat kernel signatures of two shapes are equal, the corresponding shapes or parts of the shapes are similar under isometric transformations \cite{sun2009concise}.  
Therefore, this feature is somewhat robust, making it a desirable method for point description.

\section{Point Cloud Registration Method Based on Graph Neural PDE}\label{sect:model}

In this section, we present our registration model that aims to predict the transformation between two 3D point clouds. However, point clouds may contain noise or perturbations that can compromise the robustness of the transformation prediction. Therefore, our goal is to develop a more robust method for the registration task. 

Our model is called the \textit{Point Cloud Diffusion Transformer (PointDifformer)}. First, we provide an overview of the PointDifformer framework, which is illustrated in \cref{fig:model_robT}. Then, we present the details of the modules and the loss function used in PointDifformer.

\subsection{Overview of PointDifformer}
Before introducing in detail the modules of PointDifformer, we provide an overview of its pipeline as follows. 
\begin{enumerate}
    \item Within a 3D point cloud frame, the neighborhood graph of each point consisting of its $K$ nearest neighbors is constructed. Points are regarded as the vertices in the graph. The $\calL_2$ distance between point features is used for neighbor acquisition. The initial features of the points are taken to be their 3D coordinates. The neighborhood graph for each point is an undirected complete graph. Then, graph neural PDE layers are applied to the neighborhood graph of each point to obtain a robust representation of the point. 
    \item Based on the robust feature representations, a self-cross attention module is applied to obtain an embedding containing point-level information interaction within a point cloud frame and between two point cloud frames. The heat kernel signature, as a robust feature descriptor for point clouds, is introduced into the attention module as the weights. 
    \item Using the above embeddings, an attention module is established to learn weights for points from different frames that indicate their correspondences. 
    \item Through the correspondence among points in the two point-cloud frames, the optimal transformation (including the translation and the rotation) can be estimated using optimization solution methods like the weighted SVD.
\end{enumerate}

\begin{figure}[!htb]
\centering
\includegraphics[width=\linewidth]{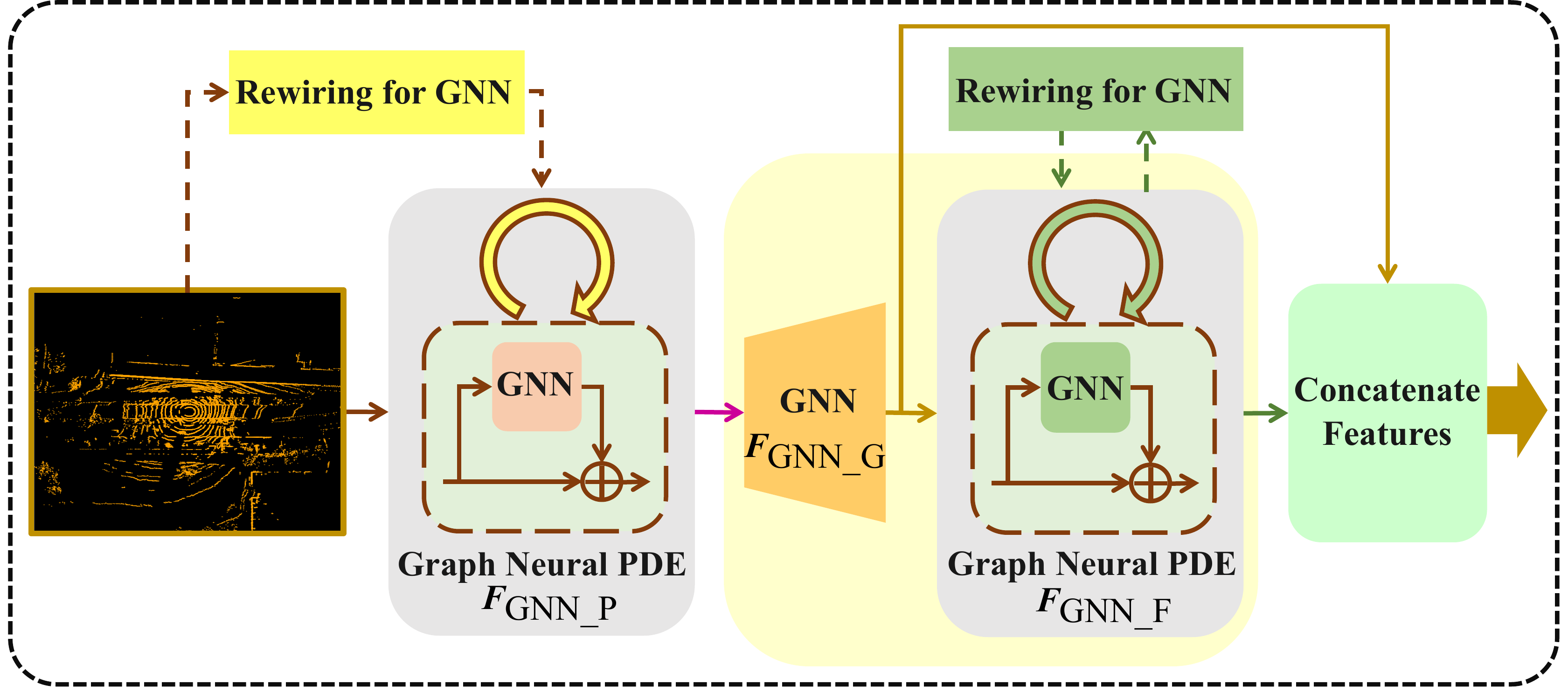}
\caption{The Point-Diffusion Net based on the graph neural PDEs for the point cloud representation. The details are provided in \cref{subsubsect:NDPCR}.}
\label{fig:model_Point_diffusion}
\end{figure}

\subsection{Model Details}\label{subsect:model_details}

\subsubsection{Point Cloud Representation with Neural Diffusion}\label{subsubsect:NDPCR} 
To represent point cloud features efficiently and robustly, we design a neural diffusion network for point cloud representation, called \textit{Point-Diffusion Net}. This module consists of GNN modules and graph neural PDE modules with different rewiring and is shown in \cref{fig:model_Point_diffusion}.
Its details are described as follows. 

\textbf{3D Points Refinement.}
To pre-process the point cloud, which has potential outliers, we first use a graph neural PDE module as a learning-based filter.
Consider a point cloud denoted by $\bX \in \mathbb{R}^{N \times 3}$. The 3-dimensional coordinates of each point are regarded as its feature map and $N$ is the number of points in the point cloud.
We construct the neighborhood graphs for the points by means of the $K$-Nearest Neighbors ($K$-NN) method using the Euclidean distance. 
Furthermore, we exploit a graph neural PDE module for feature updating, given by 
\begin{align}
\ddfrac{{\bZ}_p(t)}{t}=f_\mathrm{GNN\_P}({\bZ}_p(t)), \label{eq:GRAPH_pde_point}
\end{align}
where $f_\mathrm{GNN\_P}(\cdot)$ denotes a graph learning module and ${\bZ}_p(t)$ is the state at time $t$. 
The initial state is given by ${\bZ}_p(0)=\bX$. By integrating $f_\mathrm{GNN\_P}(\cdot)$ from $t=0$ to $t=T_p$ (using differential equation solvers \cite{ode}), we obtain the solution of \cref{eq:GRAPH_pde_point} at time $T_p$, given by
\begin{align}
{\bZ}_p(T_p) = F_\mathrm{GNN\_P}({\bZ}_p(0)) = F_\mathrm{GNN\_P}(\bX) \in \Real^{N \times 3}, \label{eq.f_x_point}
\end{align}
where $F_\mathrm{GNN\_P}(\cdot)$ can be regarded as the embedding function for the input ${\bZ}_p(0)$.
In addition, the output of this graph neural module also includes 3-dimensional features, which can also be viewed as ``generated'' points. 

\textbf{High Dimensional Feature Extraction with Graph Neural PDE.}
We extract high dimensional features for the preprocessed 3D points through a GNN module, e.g., DGCNN \cite{wang2019dynamic}. In this regard, $3$-dimensional coordinates of points are extended into $d$-dimensional features. We construct a neighborhood graph for each point using the $K$-NN method. The output from the GNN module is denoted by  
\begin{align}
{F_{\mathrm{G}}}(\bX)
&= F_{\mathrm{GNN\_G}} \circ {F_\mathrm{GNN\_P}}(\bX) \nn
&= F_{\mathrm{GNN\_G}}(\bZ_p(T_p)) \in \Real^{N \times d}, 
\end{align}
where $\circ$ denotes function composition and $F_{\mathrm{GNN\_G}}(\cdot)$ denotes the mapping of the GNN module. 
Then, we apply another graph neural PDE module to update the feature  $F_{\mathrm{G}}(\bX)$, which is described as 
\begin{align}
    \ddfrac{{\bZ}_f(t)}{t}=f_{\mathrm{GNN\_F}}({\bZ}_f(t)), \label{eq:GRAPH_pde_f}
\end{align}
where $f_{\mathrm{GNN\_F}}(\cdot)$ is also a graph learning module that deals with the neighborhood graph of 
input features. 
The initial state is given by ${\bZ}_f(0)= {F_\mathrm{G}} (\bX)$.
The equation \cref{eq:GRAPH_pde_f} is solved in the same way as that for \cref{eq:GRAPH_pde_point}. 
The output at time $T_f$ is given by 
\begin{align}
{\bZ}_f(T_f) 
& = F_\mathrm{GNN\_F}({\bZ}_f(0)) 
 = {F_\mathrm{GNN\_F}} \circ {F_{\mathrm{G}}} (\bX), 
\end{align}
where ${\bZ}_f(T_f) \in \Real^{N \times d}$, $F_\mathrm{GNN\_F}(\cdot)$ can be regarded as the embedding function for the input ${\bZ}_f(0)$.

Finally, we concatenate the output from the GNN module and the graph neural diffusion module as the eventual output for the point cloud representation module, which is denoted by
\begin{align}\label{embeddingF}
& F^{\bX} 
={F_{\mathrm{G}}}(\bX) \concat {F_\mathrm{GNN\_F}} \circ {F_{\mathrm{G}}} (\bX),
\end{align}
where $F^{\bX} \in \Real^{N \times 2d}$ and $\concat$ denotes the concatenation operation. 
Here, we follow the same approach as \cite{wang2019dynamic,wei2020end} in concatenating the feature ${\bZ}_f(T_f)={F_\mathrm{GNN\_F}} \circ {F_{\mathrm{G}}} (\bX)$ with its corresponding hidden feature $F_{\mathrm{G}}(\bX)$ to retain more information. Our numerical experiments indicate that this is a better approach than using only ${\bZ}_f(T_f)$.

\begin{figure*}[!htb]
\centering
\includegraphics[width=0.85\linewidth]{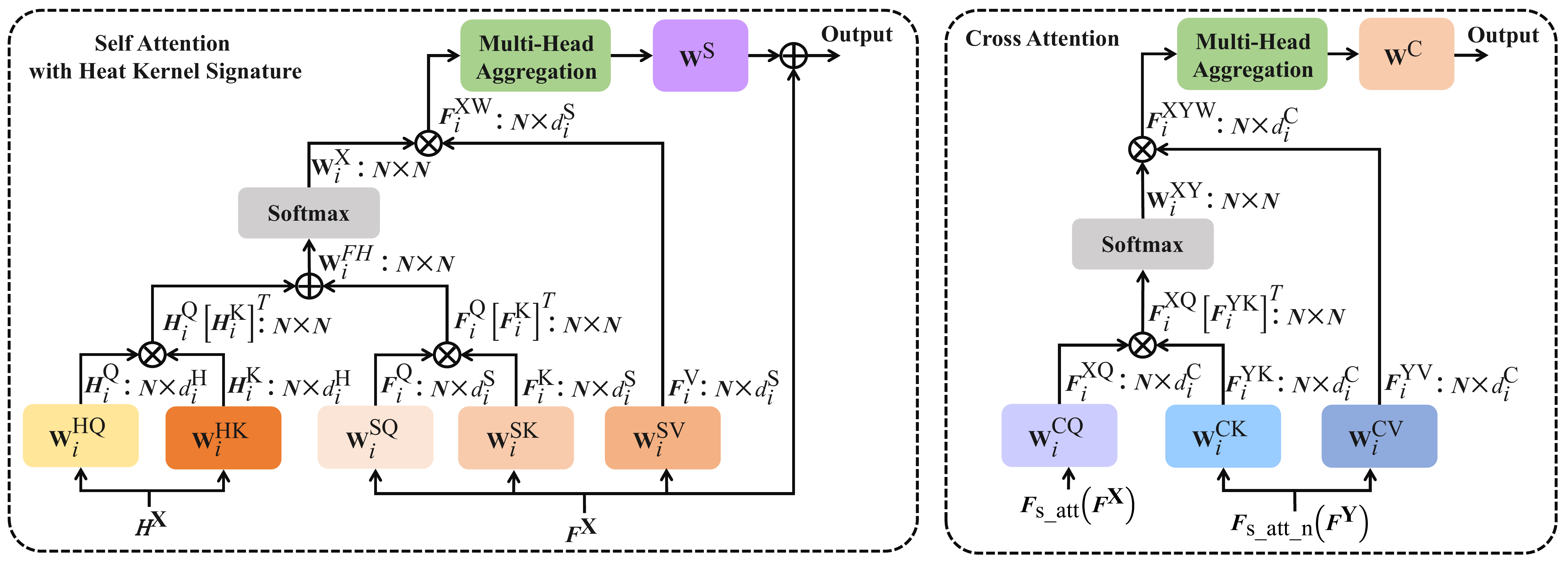}
\vspace{-0.3cm}
\caption{The modules of the self-attention with heat kernel signature and the cross-attention.}
\label{fig:architecture_attention}
\end{figure*}
\subsubsection{Self-Cross Attention Embedding Based on Heat Kernel Signature}

Based on the previous high-dimensional feature, a self-cross attention mechanism is introduced to reinforce the static structure information in each point cloud and the interactive corresponding information between a pair of point clouds, respectively. To improve the robustness of the feature extraction, we also introduce the heat kernel signature \cite{sun2009concise} into the attention mechanism as additive weights. 

For a pair of point clouds $(\bX,\bY)$, the corresponding input feature pair for the self-cross attention module is represented as $(F^{\bX}, F^{\bY})$ using the embedding from \cref{embeddingF}.
The embedding from the self-cross attention module \gls{wrt} ${\bX}$ is given by
\begin{align}
F^{\bX}_{\mathrm{sc\_att}}(F^{\bX}, F^{\bY}) =  F_{\mathrm{s\_att}}(F^{\bX}) + F^{\bX}_{\mathrm{c\_att}}(F^{\bX}, F^{\bY}), \label{eq:sc_att} 
\end{align}
where $F_{\mathrm{s\_att}}(F^{\bX})$ and $F^{\bX}_{\mathrm{c\_att}}(F^{\bX}, F^{\bY})$ are the features based on the self-attention and the cross-attention, respectively. 
The details of the features $F_{\mathrm{s\_att}}(F^{\bX})$ and $ F^{\bX}_{\mathrm{c\_att}}(F^{\bX}, F^{\bY})$ are described as follows. 

\textbf{Self-attention Feature.}
To improve the robustness, the heat kernel signature is introduced into the self-attention module.
For a point cloud pair $(\bX, \bY)$, the corresponding heat kernel signature pair is denoted by $(H^{\bX}, H^{\bY})$. 
Using the normalized $F^{\bX}$ and $H^{\bX}$ as the inputs for the self-attention module, we have 
\begin{align}
& F_{\mathrm{s\_att}}(F^{\bX}) \nn
& = {\bW}^{\text{S}} \concat\limits_{i=1}^{S_{\mathrm{head}}}
\Big\{F_{\mathrm{softmax}}\Big( \frac{(\bW^{\text{SQ}}_{i} F^{\bX}) (\bW^{\text{SK}}_{i} F^{\bX})\T}{\sqrt{d^{\text{S}}_i}} \nn
& \qquad + \frac{(\bW^{\text{HQ}}_{i}H^{\bX}) (\bW^{\text{HK}}_{i} H^{\bX})\T}{\sqrt{d^{\text{H}}_i}} \Big) (\bW^{\text{SV}}_{i} F^{\bX})\Big\} 
+ F^{\bX}, \label{eq.F_s_att_w}
\end{align}
where $(\cdot)\T$ denotes the transpose operation,
$S_{\mathrm{head}}$ is the number of multi-heads for the attention, 
$\bW^{\text{SQ}}_{i}$, $\bW^{\text{SK}}_{i}$, $\bW^{\text{SV}}_{i}$, $\bW^{\text{HQ}}_{i}$, $\bW^{\text{HK}}_{i}$, and ${\bW}^{\text{S}}$ are learnable layers, and
$d^{\text{S}}_i$ and $d^{\text{H}}_i$ are the dimensions for the point cloud features and heat kernel signatures in $i$-th attention head, respectively. 
The function $F_{\mathrm{softmax}}(\cdot)$ denotes row-wise softmax.  

The heat kernel signature is implemented as follows.
\begin{itemize}
\item \textit{Heat kernel signature acquisition.} 
We compute this feature based on the point cloud using the formula \cref{eq.heat_kernel}.
Since the function $h(x, t)$ in \cref{eq.heat_kernel} is a robust local geometric descriptor containing large-scale information \cite{chen2019lassonet}, we use it to robustly describe the repeatable features for point clouds. 

\item \textit{Embedding.}
We process the heat kernel signatures using the graph neural PDE module and the Fully Connected (FC) layer to obtain the embedding.
Similar to \cref{eq:GRAPH_pde_point}, the graph neural PDE is used as a filter for the heat kernel signatures.
The graph construction in the graph neural PDE is based on the $K$-NN for the heat kernel signatures.

\item \textit{Self-attention weights.}
After embedding the heat kernel signatures, they are input into the self-attention module as the additive weights.
By introducing extra information from the heat kernel signature, the robustness of the point cloud representation is enhanced in the self-attention module.  
\end{itemize}

\textbf{Cross-attention Feature.}
Based on the self-attention feature $F_{\mathrm{s\_att}}(F^{\bY})$ for the point cloud $\bY$, we acquire the cross-attention features for $F^{\bX}$. 
$F_{\mathrm{s\_att}}(F^{\bY})$ is input into a Feed Forward Network (FFN) \cite{vaswani2017attention} to obtain the feature
\begin{align} 
F_{\mathrm{s\_att\_n}}(F^{\bY}) = F_{\mathrm{FNN}}(F_{\mathrm{s\_att}}(F^{\bY})) + F_{\mathrm{s\_att}}(F^{\bY}), 
\end{align}
where $F_{\mathrm{FNN}}$ denotes the FFN consisting of two linear layers with normalization operation and the Rectified linear activation function (ReLU). 
Inputting normalized $F_{\mathrm{s\_att\_n}}(F^{\bY})$ and $F_{\mathrm{s\_att}}(F^{\bx})$ into the cross-attention module, we have 
\begin{align}
& F^{\bX}_{\mathrm{c\_att}}(F^{\bX}, F^{\bY}) = {\bW}^{\text{C}} \times \nn
& \concat\limits_{i=1}^{C_{\mathrm{head}}}
\Big\{F_{\mathrm{softmax}} \Big( \frac{(\bW^{\text{CQ}}_{i} F_{\mathrm{s\_att}}(F^{\bX}))(\bW^{\text{CK}}_{i} F_{\mathrm{s\_att\_n}}(F^{\bY}))\T}{\sqrt{d^{\text{C}}_i}} \Big) \nn 
& \qquad \qquad \times (\bW^{\text{CV}}_{i} F_{\mathrm{s\_att\_n}}(F^{\bY})) \Big\}, 
\end{align}
where $\bW^{\text{CQ}}_{i}$, $\bW^{\text{CK}}_{i}$, $\bW^{\text{CV}}_{i}$, and ${\bW}^{\text{C}}$ are learnable layers, $d^{\text{C}}_i$ is the feature dimension for the $i$-th attention head. 
The remaining notations are similar to those in \cref{eq.F_s_att_w}. 

The joint feature $F^{\bX}_{\mathrm{sc\_att}}(F^{\bX}, F^{\bY})$ based on the self-cross attention module is obtained as mentioned in \cref{eq:sc_att}.
Inputting $F^{\bX}_{\mathrm{sc\_att}}(F^{\bX}, F^{\bY})$ into the FNN, we have the embedding from the self-cross attention module as  
\begin{align}
F^{\bX}_{\mathrm{sc\_{\bX\bY}}} = F_{\mathrm{FNN}}(F^{\bX}_{\mathrm{sc\_att}}(F^{\bX}, F^{\bY})) + F^{\bX}_{\mathrm{sc\_att}}(F^{\bX}, F^{\bY}),
\end{align}
whose normalization is regarded as the final output of this module.

Similarly, the above self-cross attention module is also available for the point cloud $\bY$ to obtain its output $F^{\bY}_{\mathrm{sc\_{\bY\bX}}}$.
The architectures of the self-attention and the cross-attention are shown in \cref{fig:architecture_attention}.

\subsubsection{Attention-Based Keypoint Correspondence}

Using the attention mechanism, the information of point cloud $\bX$ can be involved in the embedding for the point cloud $\bY$. 
By resorting to the self-cross attention embeddings, we can obtain the weighted $\bY$ denoted by $\bY^{\bX}$ which is regarded as the transformed point cloud corresponding to the $\bX$. 
The details are given as follows. 

\textit{i)} We compute the attention weight matrix 
\begin{align}\label{eq.att_w_matrix}
    & \bW^{\mathrm{att}}
    = F_{\mathrm{softmax}} \parens*{\frac{ (F^{\bX} + F^{\bX}_{\mathrm{sc\_{\bX\bY}}}) (F^{\bY}  + F^{\bY}_{\mathrm{sc\_{\bY\bX}}})\T}{\sqrt{d^{\mathrm{att}}}} },
\end{align}
where the feature dimension $d^{\mathrm{att}}=2d$ and $F_{\mathrm{softmax}}$ denotes the row-wise softmax function. 

\textit{ii)} For each point $\bx_i$ in the point cloud $\bX$, we select its corresponding point $\by_{j^*_{i}}$ in the point cloud $\bY$, which has the highest similarity with the $\bx_i$.
Furthermore, based on the Top-$K'$ similarity scores, we select the corresponding point pairs $\set*{(\bx_i, \by_{j^*_{i}}): i=1,2,...,K'}$.
Specifically, for the $i$-th row of $\bW^{\mathrm{att}}$, we have 
\begin{align}
    j^*_{i} = \max_{j} {\bw^{\mathrm{att}}_{i,j}}, 
\end{align}
where ${\bw^{\mathrm{att}}_{i,j}}$ ($i,j \in \set*{1,2,...,N}$) denotes the element in the $i$-th row and $j$-th column among the $\bW^{\mathrm{att}}$.
Then, based on the ${\bw^{\mathrm{att}}_{i,j^*_i}}$ ($i \in \set*{1,2,...,N}$), we select the Top-$K'$ point pairs $\set*{(\bx_i, \by_{j^*_{i}}): i=1,2,...,K'}$ to obtain the updated point cloud pairs $(\bX, \bY)$, where abusing notations $\bX$ and $\bY$ are used.

\textit{iii)} Based on the updated point clouds $\bX$ and $\bY$ consisting of the Top-$K'$ points, we have the corresponding updated attention weight matrix similar to \cref{eq.att_w_matrix}, which is denoted by an abusing notation $\bW^{\mathrm{att}}$.
Then, we have the weighted $\bY$ as 
\begin{align}
    \bY^{\bX} = \bW^{\mathrm{att}} \bY,
\end{align}
whose the point number is also $K'$ the same as that in the $\bX$.  

In general, the points $\bx_i$ and $\by^{\bx}_i$ ($i=1,2,...,K'$) from the point clouds $\bX$ and $\bY^{\bX}$ are treated as the correspondence points. 

\subsubsection{Transformation Prediction}
By resorting to the correspondence of points, we can predict the transformation or relative pose between two point clouds. Consider the Mean Squared Error (MSE) given by 
\begin{align}
\ell_{\mathrm{MSE}}(\hat \bR, \hat \bt) = \frac{1}{K'} \sum_{i=1}^{K'} \norm{ \hat \bR \bx_i + \hat \bt - \by^{\bx}_i }_2,
\end{align}
in which $\| \cdot \|_2$ denotes the $\calL_2$ norm, $\bx_i = [x_i^{(1)}, x_i^{(2)}, x_i^{(3)}]\T$ and $\by^{\bx}_i = [{y^{x}_i}^{(1)}, {y^{x}_i}^{(2)}, {y^{x}_i}^{(3)}]\T$ where $x_i^{(l)}$ and ${y^{x}_i}^{(l)}$ ($l \in \{1,2,3\}$) are elements from $\bx_i$ and $\by^{\bx}_i$, respectively. The $\hat \bR \in \Real^{3 \times 3}$ and $\hat \bt \in \Real^{3 \times 1}$ are the predicted results \gls{wrt} the ground-truth rotation $\bR \in \Real^{3 \times 3}$ and translation $\bt \in \Real^{3 \times 1}$. 
The optimal results of $\hat \bR$ and $\hat \bt$ are given by
\begin{align}\label{eq.optimization_Rt}
{\hat \bR}^*, {\hat \bt}^* = \argmin_{\hat \bR, \hat \bt} \ell_{\mathrm{MSE}}(\hat \bR, \hat \bt).
\end{align}

Then, we use the weighted SVD \cite{besl1992method,qin2022geometric} to solve the optimization problem in \cref{eq.optimization_Rt}.
Specifically, the weighted mean of the $\set*{\bx_i: i=1,2,...,K'}$ and the $\set*{\by^{\bx}_i: i=1,2,...,K'}$ are first computed as 
\begin{align}
& {\bx}_{\bw} = \frac{1}{K'}\sum_{i=1}^{K'} {\bw}^{\bx}_i \bx_i, \\
& {\by}^{\bx}_{\bw} = \frac{1}{K'}\sum_{i=1}^{K'} {\bw}^{\by}_i {\by^{\bx}}_i,
\end{align}
where ${\bw}^{\bx}_i \in \Real^{3 \times 1}$ and ${\bw}^{\by}_i \in \Real^{3 \times 1}$ are trainable weights. 

Furthermore, the weighted cross-covariance matrix $\bM$ is given by
\begin{align}
& {\bM} = \sum_{i=1}^{K'} ({\bx_i}-{\bx}_{\bw})({\bw}^{\bM}_i ({\by^{\bx}_i}-{\by}^{\bx}_{\bw}))\T,
\end{align} 
where $(\cdot)\T$ denotes the transpose operation, ${\bw}^{\bM}_i \in \Real^{3 \times 1}$ is a trainable weight.

Similar to the procedure of SVD mentioned in \cite{wang2019deep,wei2020end}, the matrix $\bM$ can be decomposed as 
\begin{align}
\bM = \bU \Lambda \bV \T,
\end{align}
where $\bU$ and $\bV$ are unitary matrices and $\Lambda$ is a rectangular diagonal matrix with non-negative real diagonal elements. 
Furthermore, the transformation prediction (including the predicted rotation ${\hat \bR}^*$ and the translation ${\hat \bt}^*$) can be obtained as
\begin{align}
& {\hat \bR}^* = \bV \bU \T, \\  
& {\hat \bt}^* = -{\hat \bR}^* {\bx}_{\bw} + {\by}^{\bx}_{\bw}.
\end{align}

\subsubsection{Loss function}
As the point $\by^{\bx}_i$ corresponds to the point $\bx_i$, we use the corresponding point loss given by
\begin{align}\label{eq:Loss_p2p}
\calL_{\mathrm{point}} = \frac{1}{K'} \sum_{i=1}^{K'} \norm{{\hat \bR}^* \bx_i + {\hat \bt}^* - \by^{\bx}_i}_2. 
\end{align}
To quantify the deviation between the ground truth and the predicted results \gls{wrt} rotation and translation, we use the loss given by 
\begin{align}\label{eq:Loss_rt}
\calL_{\mathrm{rt}} 
& = \exp{(-\gamma_{\mathrm{t}})}\norm{{\hat \bt}^*-\bt}_2 + \gamma_{\mathrm{t}}  \nn
& \quad + \exp{(-\gamma_{\mathrm{r}})}\norm{\bR\T{\hat \bR}^*-\bI}_2 + \gamma_{\mathrm{r}}, 
\end{align}
where $\bI$ denotes the identity matrix, and $\gamma_{\mathrm{t}}$ and $\gamma_{\mathrm{r}}$ are learnable parameters.
The learnable weights based on $\gamma_{\mathrm{t}}$ and $\gamma_{\mathrm{r}}$ are inpsired by the loss in \cite{kendall2017geometric,Wang2022robustloc}. 
The total loss combining $\calL_{\mathrm{point}}$ and $\calL_{\mathrm{rt}}$ is given by 
\begin{align}\label{eq:Loss_total}
& \calL_{\mathrm{total}} = \exp{(-\eta_{\mathrm{p}})}\calL_{\mathrm{point}}  + \eta_{\mathrm{p}}  + \exp{(-\eta_{\mathrm{rt}})}\calL_{\mathrm{rt}}  + \eta_{\mathrm{rt}}, 
\end{align}
where $\eta_{\mathrm{p}}$ and $\eta_{\mathrm{rt}}$ are learnable parameters.

\section{Experiments}\label{sect:exper}

\subsection{Dataset preparation}\label{sect:data-preparation}

\textbf{vReLoc Dataset.}
The vReLoc Dataset is a publicly available indoor dataset,\footnote{https://github.com/loveoxford/vReLoc} containing LiDAR point clouds and camera images. 
In this paper, we randomly generate a transformation matrix for each point cloud to obtain a pair of point cloud frames. The transformation matrix is based on translation along the $x$, $y$, and $z$ axes, as well as rotation along the roll, pitch, and yaw axes. The generated transformation matrix is regarded as the ground truth.
The generated translation values are uniformly sampled from the intervals $[-1, 1]$, $[-2, 2]$, and $[-0.5, 0.5]$ along the $x$-, $y$-, and $z$-axes, respectively. The generated rotation values are uniformly sampled from the intervals $[0^\circ, 5^\circ]$, $[0^\circ, 5^\circ]$, and $[0^\circ, 30^\circ]$ around the roll, pitch, and yaw axes, respectively.

\textbf{Boreas Dataset.}
The Boreas dataset is a publicly available outdoor dataset\footnote{https://www.boreas.utias.utoronto.ca/} that comprises multi-sensor data, including LiDAR and camera data. It presents various environmental scenarios, such as sunny, night, and rainy conditions, as it was collected over the course of one year by repeatedly driving a specific route. Furthermore, the dataset provides post-processed ground-truth poses with centimeter-level accuracy, which offers the transformation matrix required for two consecutive LiDAR point clouds. 
The Boreas datasets undergo preprocessing involving distortion correction of LiDAR point clouds, as detailed in \cite{burnett2023boreas}. However, these preprocessing techniques do not completely eliminate all distortions in LiDAR point clouds, such as the tailing phenomenon \cite{song2023registration}. Additionally, noise can still be present in environments with adverse weather conditions, dynamic objects or vehicles, and pedestrians, which can affect the accuracy of the LiDAR measurements.

\textbf{KITTI Dataset.}
The KITTI dataset is a publicly available outdoor dataset\footnote{http://www.cvlibs.net/datasets/kitti/} that provides multi-sensor data for autonomous driving. It includes LiDAR point clouds of street scenes captured using the Velodyne Laserscanner in Karlsruhe, Germany, with tens of thousands of LiDAR points in each frame. The dataset consists of $11$ sequences (from sequence ``$0$'' to ``$10$'') depicting different street scenes, and global ground-truth poses are available for each sequence.
Similar to the Boreas Dataset, we can use the ground-truth poses to obtain the transformation matrix between each pair of adjacent LiDAR point clouds in the KITTI dataset.
While the KITTI dataset incorporates preprocessing for point cloud calibration \cite{geiger2012we} through auxiliary sensors, such as Global Positioning System (GPS) and Inertial Measurement Unit (IMU), perturbations similar to those in the Boreas datasets persist.

\subsection{Experimental Details}

\textbf{Model Setting.}
We set $d=256$ in \cref{eq:GRAPH_pde_point}.
To deal with the neighborhood graph of the $K$ nearest neighbors ($K=20$), we set the GNN layer $f_\mathrm{GNN\_P}$ in the graph neural PDE \cref{eq:GRAPH_pde_point} to be the union EdgeConv layers \cite{wang2019dynamic} which is also regarded as a kind of DGCNN.
There are $5$ EdgeConv layers used in the DGCNN block, whose hidden input and output dimensions are given by $[6,16]$, $[16,16]$, $[16,32]$, $[32,64]$ and $[128,3]$, respectively. 
We set the graph learning module for the $F_\mathrm{GNN\_G}$ to another DGCNN, in which $5$ EdgeConv layers are used with the hidden input and output dimensions $[6,64]$, $[64,64]$, $[64,128]$, $[128,256]$ and $[512,256]$, respectively. 
When using the DGCNN-based graph neural PDE for $F_\mathrm{GNN\_F}$, there are $2$ EdgeConv layers used in the DGCNN block with the hidden input and output dimensions $[256,256]$ and $[768,256]$.
As for the self-cross attention module, there are $4$ attention heads with $128$ hidden features for each attention head, which implies $512$ hidden features in total.  
We adopt the Adam optimizer \cite{kingma2014adam} in the training, where the learning rate is set as $0.0001$. 
We set the number of training epochs as $50$.

\textbf{Baseline Methods.}
To demonstrate the superior performance of PointDifformer, we compared it against several baseline methods, including ICP \cite{segal2009generalized}, DCP \cite{wang2019deep}, HGNN \cite{feng2019hypergraph}, VCR-Net \cite{wei2020end}, PCT \cite{guo2021pct}, and GeoTransformer \cite{qin2022geometric}. 
ICP is an iterative optimization method that does not require neural networks for feature learning, meaning that it does not need a training process. On the other hand, the other methods utilize learned point cloud features to determine point correspondence such as DCP, VCR-Net and GeoTransformere. We further enhanced HGNN and PCT with attention modules for point correspondence registration, which we refer to as HGNN++ and PCT++, respectively. 

\begin{figure*}[!htb]
\centering
\includegraphics[width=0.98\linewidth]{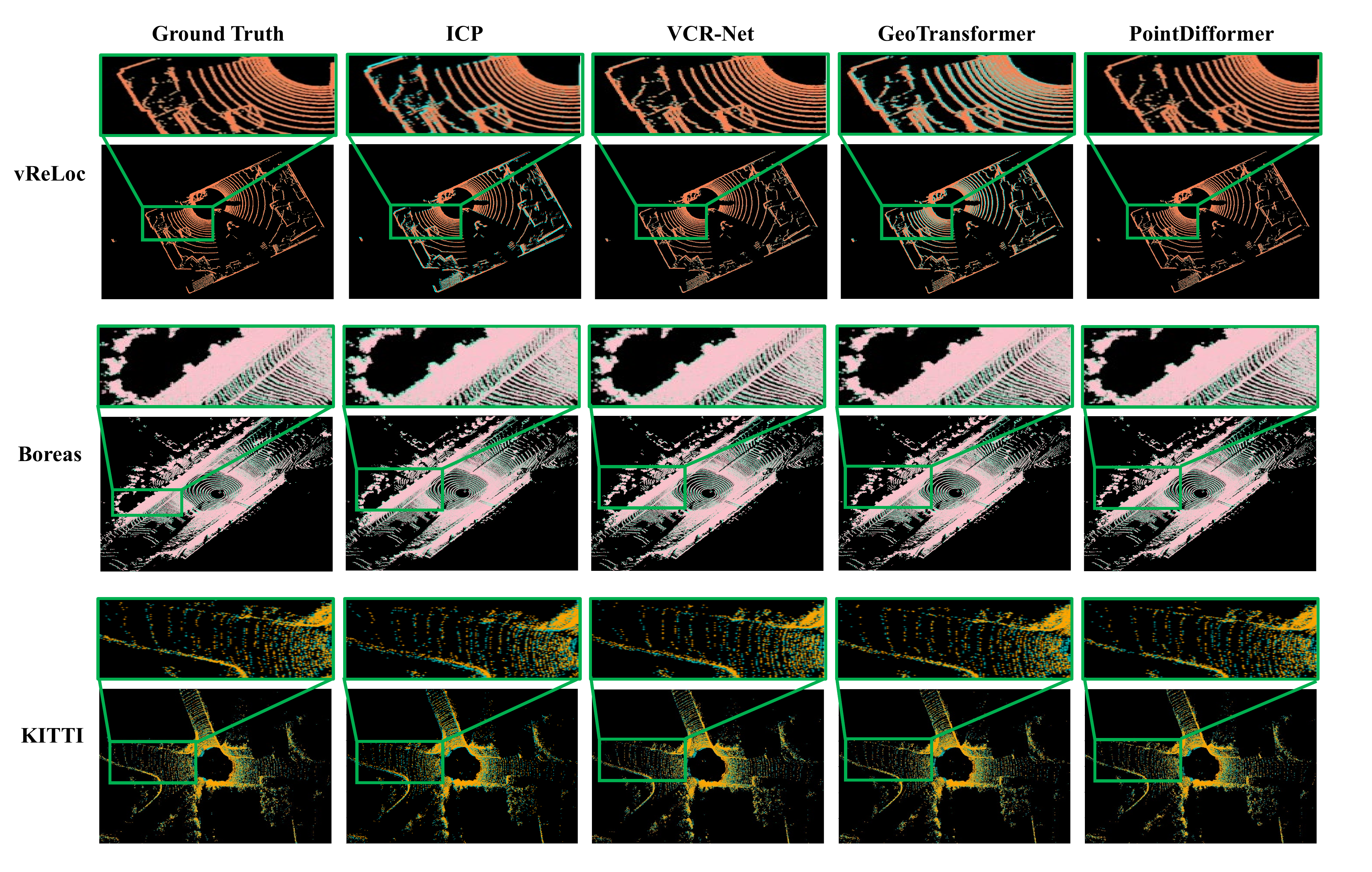}
\caption{Examples of point cloud frame pairs transformed using different prediction methods to align the second frame with the coordinate system of the first frame.}
\label{fig:combine_T_example}
\end{figure*}

\subsection{Point Cloud Registration Performance}

\subsubsection{Evaluation on Indoor vReLoc Dataset}
To compare our method with other baselines, we evaluate them on their ability to predict transformations between two nearby point cloud frames from the vReLoc dataset. For training, we use sequences ``$3$'', ``$6$'', and ``$9$'', while for testing, we use sequences ``$14$'' and ``$16$''. We evaluate the performance of these methods using statistics such as Mean Absolute Error (MAE) and Root Mean Square Error (RMSE) for the predicted relative translation and rotation results.
We also utilize Registration Recall (RR) as a metric, which is defined in \cite{yu2023rotation, qin2022geometric, zhang20233d}. RR measures the percentage of point cloud frames that achieve a certain threshold of registration accuracy. In our evaluation, we use a fine-tuned threshold for performance comparison.

\begin{figure}[!htb]
\centering
\includegraphics[width=\linewidth]{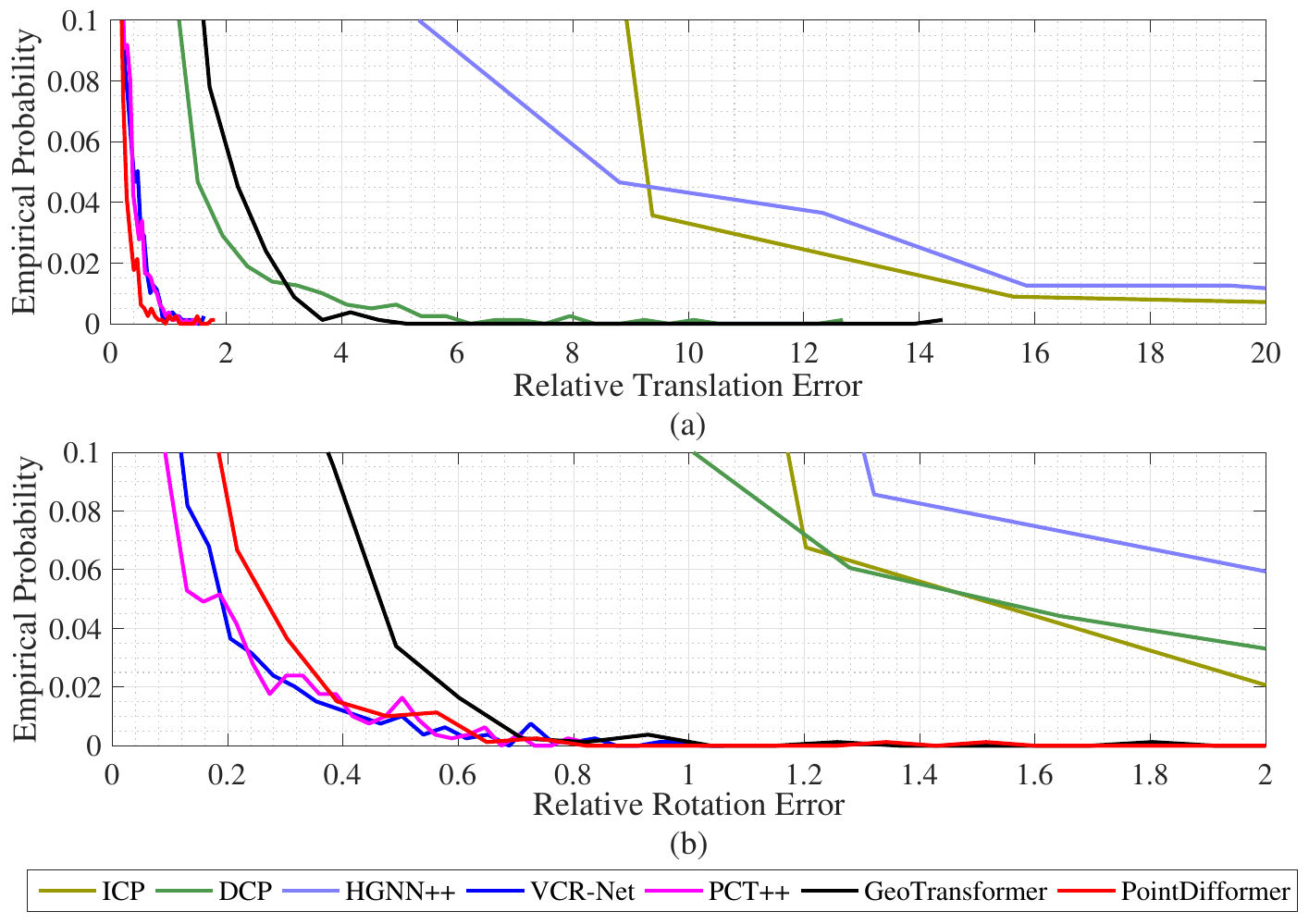}
\vspace{-0.3cm}
\caption{The empirical probability of relative translation (centimeter [cm]) and rotation (degree [$^{\circ}$]) errors on the vReLoc dataset.
}
\label{fig:error-vreloc}
\end{figure}

From \cref{tab:vReLoc_Rt_clear}, we observe that PointDifformer outperforms the other baselines without neural diffusions in terms of relative translation and rotation prediction. This suggests that the graph neural PDE modules play a positive role in point cloud registration.
Further analysis of PointDifformer in \cref{fig:error-vreloc} shows that the translation and rotation errors lie in a small region close to zero. 
In \figref{fig:error-vreloc}, the empirical probability or the relative frequency of an error value, is the ratio of the number of errors within a small bin around the error value to the total number of trials.
The relative translation error and rotation error in \cref{fig:error-vreloc} are calculated as the absolute values of the difference between the corresponding predictions and the ground truth. 
Using the predicted transformation between two point cloud frames, we can transform the second frame into the coordinate system of the first frame to achieve alignment of the point clouds.
We show several examples of point cloud alignment based on the predicted transformation in \cref{fig:combine_T_example}, where the degree of overlap between the two frames increases with the accuracy of the predicted transformation.

\begin{table}[!htp]
\caption{Performance of point cloud registration prediction on the vReLoc dataset. The best and second-best results under different metrics are highlighted in \textbf{bold} and \underline{underlined}, respectively.}
\label{tab:vReLoc_Rt_clear}
\centering
\newcommand{\tabincell}[2]{\begin{tabular}{@{}#1@{}}#2\end{tabular}}
\begin{tabular}{c|c c c c c} 
\hline\hline
\multirow{2}{*}{Method}         & \multicolumn{2}{c}{\tabincell{c}{Relative Translation \\ Error (centimeter [cm])}}  &  \multicolumn{2}{c}{\tabincell{c}{Relative Rotation \\ Error (degree [$^\circ$])}} &  \multirow{2}{*}{\tabincell{c}{RR \\ (\%) }}\\
                                &  \textit{MAE}   & \textit{RMSE}   & \textit{MAE} &   \textit{RMSE}   &  \\
\hline
ICP               & 2.20               & 12.69                 & 0.20              & 1.35              & 96.2            \\ \hline
DCP               & 1.35               & 3.26                  & 0.45              & 1.31              & 85.4            \\ \hline
HGNN++            & 3.33               & 11.29                 & 0.36              & 1.57              & 83.5            \\ \hline
VCR-Net           & \underline{0.25}   & 0.45                  & \underline{0.04}  & \underline{0.11}  & \textbf{99.9}   \\ \hline
PCT++             & \underline{0.25}   & \underline{0.44}      & 0.05              & 0.12              & \textbf{99.9}   \\ \hline
GeoTransformer    & 0.66               & 1.10                  & 0.07              & 0.16              & \textbf{99.9}   \\ \hline
PointDifformer    & \textbf{0.14}      & \textbf{0.40}         & \textbf{0.03}     & \textbf{0.10}     & \textbf{99.9}   \\
\hline\hline
\end{tabular}
\end{table}

\subsubsection{Evaluation on Outdoor Boreas Dataset}\label{sect:cross_Boreas}
\begin{figure}[!htb]
\centering
\includegraphics[width=\linewidth]{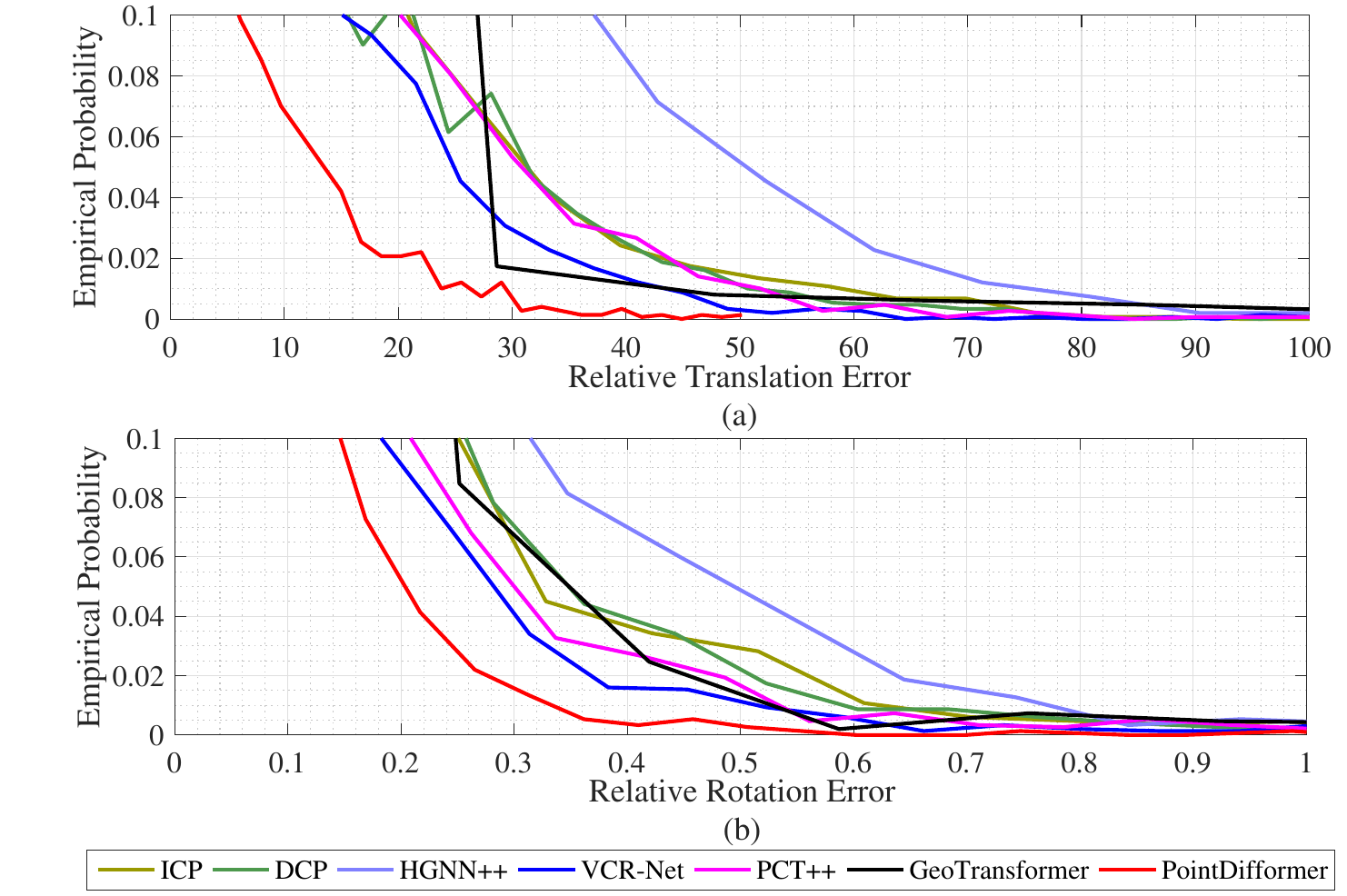}
\vspace{-0.3cm}
\caption{The empirical probability of relative translation (centimeter [cm]) and rotation (degree [$^{\circ}$]) errors on the Boreas dataset. }
\label{fig:error-boreas}
\end{figure}

We compare PointDifformer with other baselines on the outdoor Boreas dataset, where the training dataset is collected under sunny weather, and the test dataset is collected under night weather. As shown in \cref{tab:Boreas_Rt_clear}, PointDifformer outperforms the other baselines under all criteria, except for the relative translation MAE, for which GeoTransformer is slightly better. However, \cref{fig:error-boreas} shows that GeoTransformer has longer probability tails in the translation and rotation errors than PointDifformer, indicating that our method is more robust. We also present several examples of point cloud alignment using the predicted transformations in \cref{fig:combine_T_example}.
\begin{table}[!htp]
\caption{Point cloud registration performance on the Boreas dataset.}
\label{tab:Boreas_Rt_clear}
\centering
\newcommand{\tabincell}[2]{\begin{tabular}{@{}#1@{}}#2\end{tabular}}
\begin{tabular}{c|c c c c c} 
\hline\hline
\multirow{2}{*}{Method}         & \multicolumn{2}{c}{\tabincell{c}{Relative Translation \\ Error (centimeter [cm])}}  &  \multicolumn{2}{c}{\tabincell{c}{Relative Rotation \\ Error (degree [$^\circ$])}} & \multirow{2}{*}{\tabincell{c}{RR \\ (\%) }} \\
                                &  \textit{MAE}     & \textit{RMSE}  & \textit{MAE} & \textit{RMSE}           &     \\
\hline
ICP               & 10.83               & 18.28                   & 0.11                & 0.21                & 75.4    \\ \hline
DCP               & 11.63               & 17.36                   & 0.12                & 0.21                & 70.5    \\ \hline
HGNN++            & 14.41               & 23.16                   & 0.14                & 0.25                & 56.1    \\ \hline
VCR-Net           & 8.71                & \underline{13.56}       & 0.10                & \underline{0.17}    & 84.7    \\ \hline
PCT++             & 9.81                & 15.77                   & 0.10                & 0.19                & 79.6    \\ \hline
GeoTransformer    & \textbf{4.58}       & 15.78                   & \underline{0.08}    & 0.22                & \underline{94.9}    \\ \hline
PointDifformer    & \underline{6.12}    & \textbf{8.84}           & \textbf{0.07}       & \textbf{0.12}       & \textbf{96.1}    \\
\hline\hline
\end{tabular}
\end{table}

\subsubsection{Evaluation on Outdoor KITTI Dataset}\label{sect:eval_KITTI}
\begin{figure}[!htb]
\centering
\includegraphics[width=\linewidth]{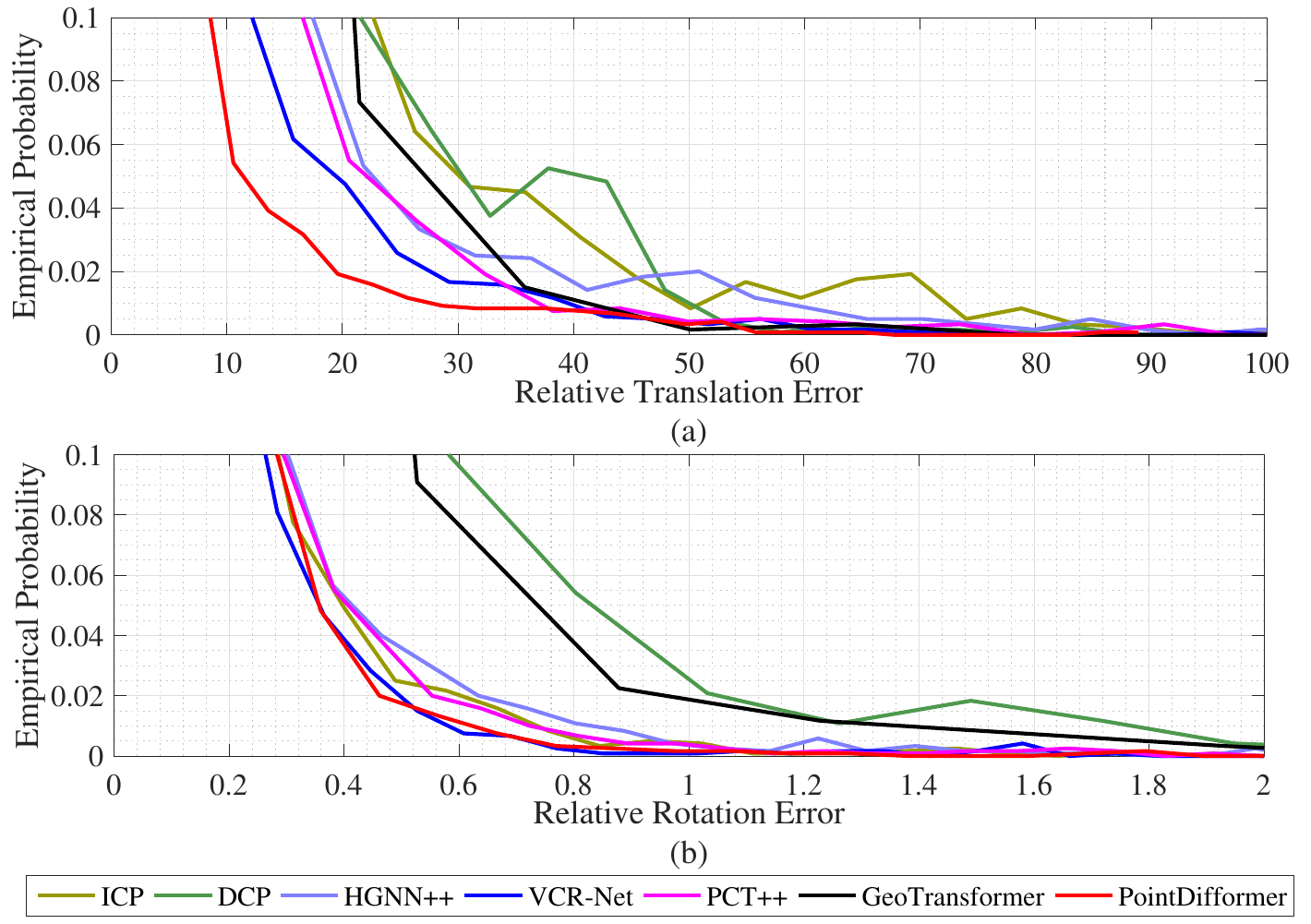}
\vspace{-0.3cm}
\caption{The empirical probability of relative translation (centimeter [cm]) and rotation (degree [$^{\circ}$]) errors on the KITTI dataset. }
\label{fig:error-kitti}
\end{figure}

We conduct point cloud registration methods on the KITTI dataset, selecting around $1600$ and $1200$ point cloud pairs for training and testing, respectively. From \cref{tab:KITTI_Rt_clear} and \cref{fig:error-kitti}, we observe that PointDifformer demonstrates superior performance compared to other baselines on the KITTI dataset, with shorter tails of relative rotation and translation error probabilities. This is similar to its performance on the Boreas dataset. 
We also present several examples of our results in \cref{fig:combine_T_example}. 
Furthermore, we train on sequences ``$0$'' to ``$8$'' and test on sequences ``$9$'' to ``$10$''. 
We observe that PointDifformer surpasses other baselines when the size of the training data is larger, as shown in \cref{tab:KITTI_Rt}. The LiDAR point clouds in the KITTI dataset have practical noise due to dynamic objects and complex environments. However, the graph neural PDE modules in PointDifformer exhibit robustness to input perturbations, as demonstrated in \cite{she2023robustmat}. This may be the reason why PointDifformer achieves more accurate predicted results when there is more practical noise in larger-sized data. 
\begin{table}[!htp]
\caption{Performance of point cloud registration prediction on the KITTI dataset.}  
\label{tab:KITTI_Rt_clear}
\centering
\newcommand{\tabincell}[2]{\begin{tabular}{@{}#1@{}}#2\end{tabular}}
\begin{tabular}{c|c c c c c} 
\hline\hline
\multirow{2}{*}{Method}         & \multicolumn{2}{c}{\tabincell{c}{Relative Translation \\ Error (centimeter [cm])}}  &  \multicolumn{2}{c}{\tabincell{c}{Relative Rotation \\ Error (degree [$^\circ$])}}  & \multirow{2}{*}{\tabincell{c}{RR \\ (\%) }} \\
                                &  \textit{MAE}   & \textit{RMSE}   & \textit{MAE} &   \textit{RMSE}          & \\
\hline
ICP               & 9.86                & 19.48                   & 0.17                & 0.27                & 87.9 \\ \hline
DCP               & 9.28                & 15.34                   & 0.26                & 0.49                & 95.0 \\ \hline
HGNN++            & 8.86                & 17.20                   & 0.20                & 0.31                & 89.9 \\ \hline
VCR-Net           & 5.31                & \underline{11.07}       & \underline{0.16}    & \underline{0.24}    & 97.3 \\ \hline
PCT++             & 6.16                & 13.96                   & 0.18                & 0.28                & 95.4 \\ \hline
GeoTransformer    & \textbf{3.93}       & 13.50                   & 0.18                & 0.50                & \textbf{97.8} \\ \hline
PointDifformer    & \underline{4.14}    & \textbf{8.86}           & \textbf{0.14}       & \textbf{0.23}       & \underline{97.7} \\
\hline\hline
\end{tabular}
\end{table}
\begin{table}[!htp]
\caption{Performance of point cloud registration prediction on the KITTI dataset with sequence ``$0$'' to ``$8$'' for training and sequence ``$9$'' to ``$10$'' for the test.}
\label{tab:KITTI_Rt}
\centering
\newcommand{\tabincell}[2]{\begin{tabular}{@{}#1@{}}#2\end{tabular}}
\begin{tabular}{c|c c c c c} 
\hline\hline
\multirow{2}{*}{Method}         & \multicolumn{2}{c}{\tabincell{c}{Relative Translation \\ Error (centimeter [cm])}}  &  \multicolumn{2}{c}{\tabincell{c}{Relative Rotation \\ Error (degree [$^\circ$])}} & \multirow{2}{*}{\tabincell{c}{RR \\ (\%)}} \\
                                &  \textit{MAE}   & \textit{RMSE}  & \textit{MAE} & \textit{RMSE}                   & \\
\hline
VCR-Net          & 7.17                 & \underline{10.88}         & 0.21                  & \underline{0.36}      & 97.2 \\ \hline
GeoTransformer   & \underline{3.45}     & 12.91                     & \underline{0.14}      & 0.76                  & \textbf{99.0} \\ \hline
PointDifformer   & \textbf{3.10}        & \textbf{5.93}             & \textbf{0.11}         & \textbf{0.17}         & \textbf{99.0} \\
\hline\hline
\end{tabular}
\end{table}

\subsubsection{Generalization from KITTI Dataset to Boreas dataset}
To cross-validate, we conduct point cloud registration by training the models on the KITTI dataset and evaluating them on the Boreas dataset. Specifically, we train the models on sequence ``$9$'' of the KITTI dataset and test them on the sequence ``night'' of the Boreas dataset. From \cref{tab:KITTI_Boreas_Rt_cross}, we observe that PointDifformer has competitive performance compared with the current state-of-the-art.
\begin{table}[!htp]
\caption{The performance of point cloud registration on the Boreas dataset for testing (using the model pre-trained on the KITTI dataset).}
\label{tab:KITTI_Boreas_Rt_cross}
\centering
\newcommand{\tabincell}[2]{\begin{tabular}{@{}#1@{}}#2\end{tabular}}
\begin{tabular}{c|c c c c c} 
\hline\hline
\multirow{2}{*}{Method}         & \multicolumn{2}{c}{\tabincell{c}{Relative Translation \\ Error (centimeter [cm])}}  &  \multicolumn{2}{c}{\tabincell{c}{Relative Rotation \\ Error (degree [$^\circ$])}}  & \multirow{2}{*}{\tabincell{c}{RR \\ (\%) }} \\
                                &  \textit{MAE}     & \textit{RMSE}  & \textit{MAE} & \textit{RMSE}           & \\
\hline
HGNN++            & 16.06               & 25.86                   & 0.15                & 0.27                & 49.8 \\ \hline
VCR-Net           & 11.97               & 19.78                   & 0.11                & \underline{0.19}    & 68.6 \\ \hline
PCT++             & 11.61               & \underline{19.57}       & 0.13                & 0.31                & 72.4 \\ \hline
GeoTransformer    & \textbf{5.97}       & 27.90                   & \underline{0.09}    & 0.33                & \underline{93.1} \\ \hline
PointDifformer    & \underline{6.63}    & \textbf{10.07}          & \textbf{0.08}       & \textbf{0.14}       & \textbf{93.3} \\
\hline\hline
\end{tabular}
\end{table}

\subsection{Robutness Evaluation}\label{sect:Robutness}

\subsubsection{Robustness against synthetic noise on the KITTI Dataset}
We investigate the robustness of PointDifformer under various types of synthetic noise, as discussed below.

\begin{table}[!htp]
\caption{Point cloud registration performance on the KITTI dataset with the Gaussian noise that follows $\calN(0, \sigma = 0.25)$. 
}
\label{tab:KITTI_Rt_noise}
\centering
\newcommand{\tabincell}[2]{\begin{tabular}{@{}#1@{}}#2\end{tabular}}
\begin{tabular}{c|c c c c c} 
\hline\hline
\multirow{2}{*}{Method}         & \multicolumn{2}{c}{\tabincell{c}{Relative Translation \\ Error (centimeter [cm])}}  &  \multicolumn{2}{c}{\tabincell{c}{Relative Rotation \\ Error (degree [$^\circ$])}} & \multirow{2}{*}{\tabincell{c}{RR \\ (\%) }} \\
                                &  \textit{MAE}     & \textit{RMSE}  & \textit{MAE} & \textit{RMSE}              & \\
\hline
ICP               & 14.97               & 26.09                   & 0.20                & 0.32                   & 69.3 \\ \hline
DCP               & 9.97                & 15.84                   & 0.29                & 0.52                   & 94.3 \\ \hline
HGNN++            & 10.62               & 18.76                   & 0.22                & 0.34                   & 88.9 \\ \hline
VCR-Net           & 6.40                & \underline{12.40}       & \underline{0.18}    & \underline{0.27}       & 96.3 \\ \hline
PCT++             & 6.85                & 14.03                   & 0.20                & 0.30                   & 95.3 \\ \hline
GeoTransformer    & \underline{5.37}    & 14.43                   & 0.25                & 0.50                   & \underline{97.5} \\ \hline
PointDifformer    & \textbf{5.23}       & \textbf{9.00}           & \textbf{0.17}       & \textbf{0.25}          & \textbf{97.7} \\
\hline\hline
\end{tabular}
\end{table}
\begin{figure}[!htb]
\centering
\includegraphics[width=\linewidth]{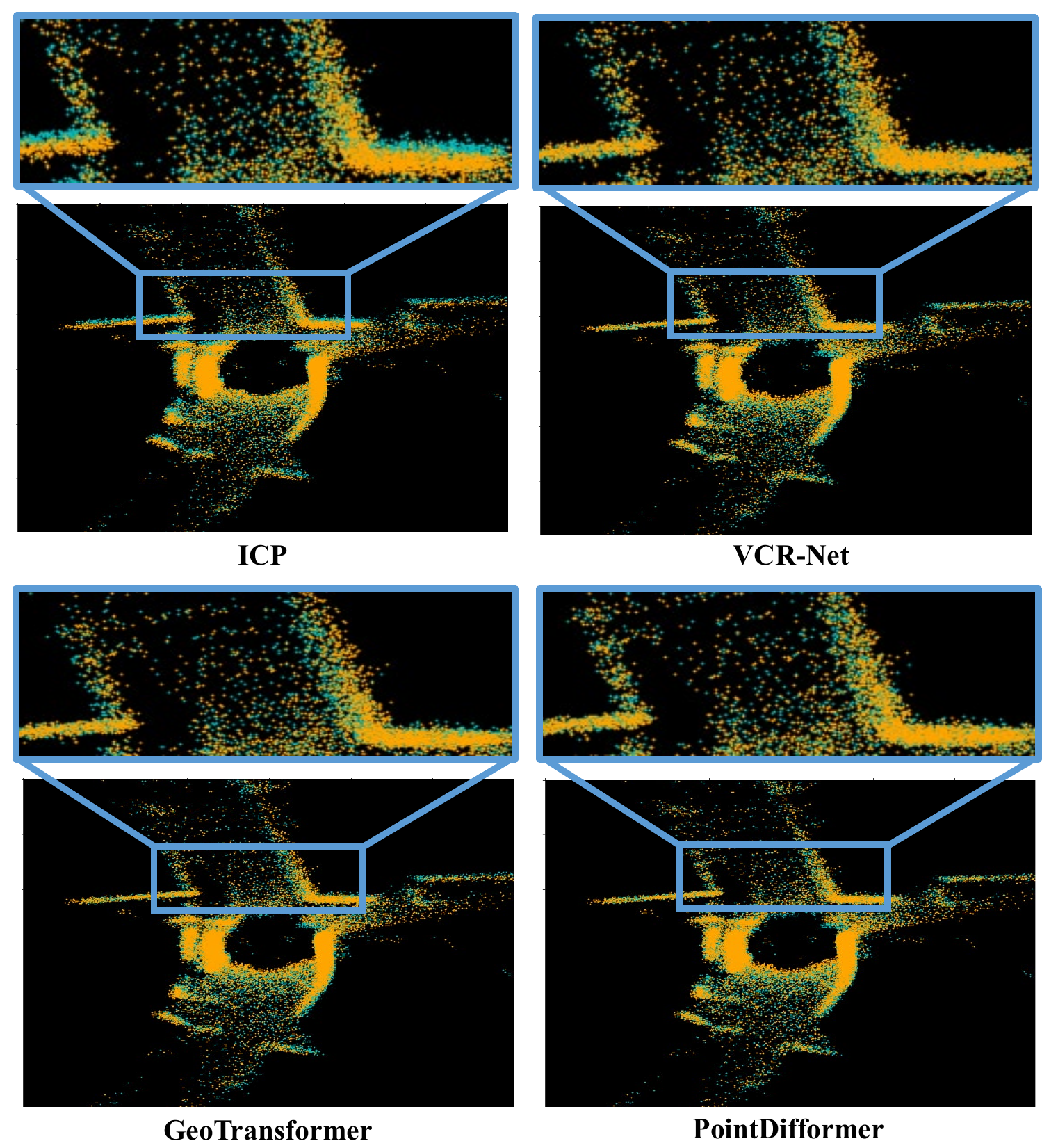}
\caption{Examples for the point cloud frame pairs from the KITTI dataset with Gaussian noise using different transformation prediction methods for alignment.}
\label{fig:kitti_T_example_Gaussian}
\end{figure}
\begin{table}[!htp]
\caption{The MAE of the predicted relative translation/rotation (centimeter [cm] / degree [$^\circ$]) on different noise powers.}
\label{tab:KITTI_noise}
\centering
\resizebox{0.48\textwidth}{!}{\setlength{\tabcolsep}{1pt} 
\begin{tabular}{c|c|c|c|c} 
\hline\hline
Metric & Methods          & $\calN(0, \sigma = 0.5)$  &  $\calN(0, \sigma = 0.75)$   & $\calN(0, \sigma = 1.0)$   \\
\hline
\multirow{3}{*}{MAE} 
&   VCR-Net              &  7.97 / \textbf{0.23}                & \underline{9.70} /  \underline{0.29} &  \underline{11.62} /  \underline{0.36} \\
&   GeoTransformer       & \underline{7.95} / 0.34              & 10.04 / 0.44                          & 13.27 / 0.54        \\ 
&   PointDifformer       &  \textbf{7.18} / \textbf{0.23}       & \textbf{8.83} / \textbf{0.27}         &  \textbf{10.38} / \textbf{0.31} \\ 
\hline
\multirow{3}{*}{RMSE} 
&   VCR-Net               &  \underline{14.08} /  \underline{0.35}    &  15.92 /  \underline{0.44}          &  \underline{17.95} /  \underline{0.55} \\
&   GeoTransformer        & 14.40 / 0.68                               & \textbf{15.00} / 0.78           & 20.15 / 0.90               \\ 
&   PointDifformer        &  \textbf{12.90} / \textbf{0.34}           & \underline{15.08} / \textbf{0.40}      & \textbf{17.31} / \textbf{0.45} \\ 
\hline\hline
\end{tabular}}
\end{table}
\begin{table}[!htp]
\caption{Point cloud registration performance on the KITTI dataset with 3D shape perturbations.}
\label{tab:KITTI_Rt_noise_3d_shape}
\centering
\newcommand{\tabincell}[2]{\begin{tabular}{@{}#1@{}}#2\end{tabular}}
\begin{tabular}{c|c c c c c } 
\hline\hline
\multirow{2}{*}{Method}         & \multicolumn{2}{c}{\tabincell{c}{Relative Translation \\ Error (centimeter [cm])}}  &  \multicolumn{2}{c}{\tabincell{c}{Relative Rotation \\ Error (degree [$^\circ$])}} & \multirow{2}{*}{\tabincell{c}{RR \\ (\%) }}\\
                                &  \textit{MAE}  & \textit{RMSE}& \textit{MAE} &   \textit{RMSE}        & \\
\hline
ICP               & 12.60   & 24.92                           & \underline{0.18}  & \underline{0.32}    & 80.4 \\ \hline
DCP               & 12.63   & 23.04                           & 0.28              & 0.48                & 85.8 \\ \hline
HGNN++            & 11.39            & 22.27                  & 0.21              & 0.36                & 85.1 \\ \hline
VCR-Net           & 6.39             & 13.69                  & \underline{0.18}  & 0.36                & 95.4 \\ \hline
PCT++             & 7.23             & 16.23                  & 0.21              & 0.36                & 93.6  \\ \hline
GeoTransformer    & \textbf{3.89}    & \underline{13.08}      & \underline{0.18}  & 0.44                & \textbf{97.8} \\ \hline
PointDifformer    & \underline{4.14} & \textbf{8.99}          & \textbf{0.14}     & \textbf{0.22}       & \textbf{97.8} \\
\hline\hline
\end{tabular}
\end{table}
\begin{figure}[!htb]
\centering
\includegraphics[width=\linewidth]{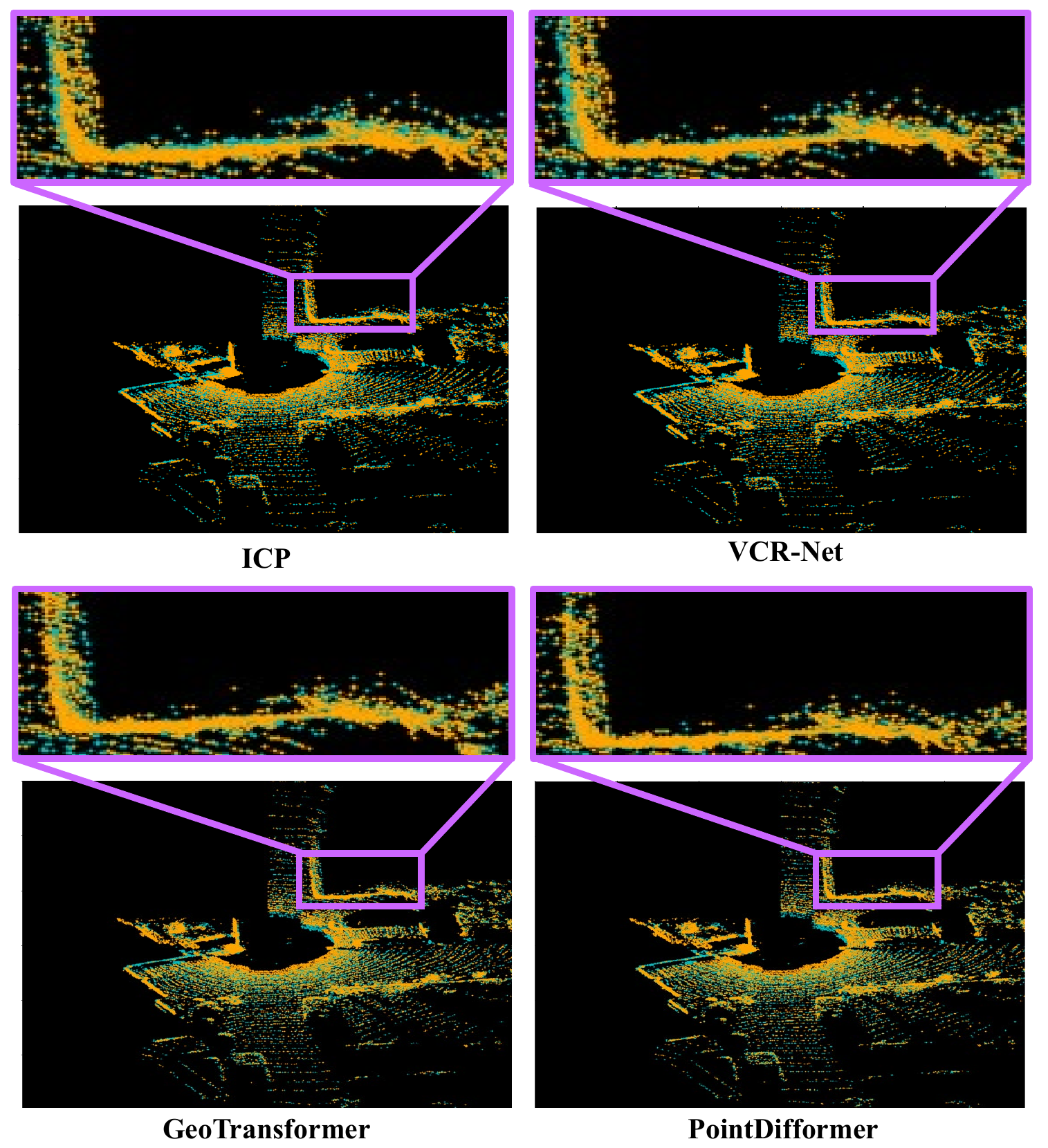}
\caption{Examples for noisy point cloud frames with 3D shape perturbations using different transformation prediction methods for alignment.}
\label{fig:kitti_T_example_3d_shape}
\end{figure}

\textbf{Performance on the KITTI dataset with Gaussian noise.}
To evaluate the robustness of PointDifformer, we add white Gaussian noise $\calN(0,\sigma)$ to the original KITTI dataset, similar to the experiments for \cref{tab:KITTI_Rt_clear}. Based on the results presented in \cref{tab:KITTI_Rt_noise}, we observe that PointDifformer surpasses the other benchmark methods in terms of relative translation and rotation errors. Additionally, we present examples of point cloud alignment using the predicted transformation in \cref{fig:kitti_T_example_Gaussian}.
We also evaluate the robustness of our method and other baselines to different noise powers. As shown in \cref{tab:KITTI_noise}, PointDifformer demonstrates superior robustness \gls{wrt} additive Gaussian noise compared to the other baselines across different noise powers.

\textbf{Performance on the noisy KITTI dataset with 3D shape perturbations.}
We next introduce 3D shape perturbations to the original KITTI dataset. This is achieved by removing certain parts of the original point clouds, resulting in an imperfect 3D shape. 
Specifically, we remove a $25$ m $\times 15$ m region in the lower left corner of each point cloud frame measuring $60$ m $\times 30$ m.
The results presented in \cref{tab:KITTI_Rt_noise_3d_shape} show that PointDifformer still outperforms the baselines in terms of both criteria for relative rotation error and the RMSE for relative translation prediction. Some examples are presented in \cref{fig:kitti_T_example_3d_shape}.

\subsubsection{Robustness against natural noise on the Boreas Dataset}
\begin{figure}[!htb]
\centering
\includegraphics[width=\linewidth]{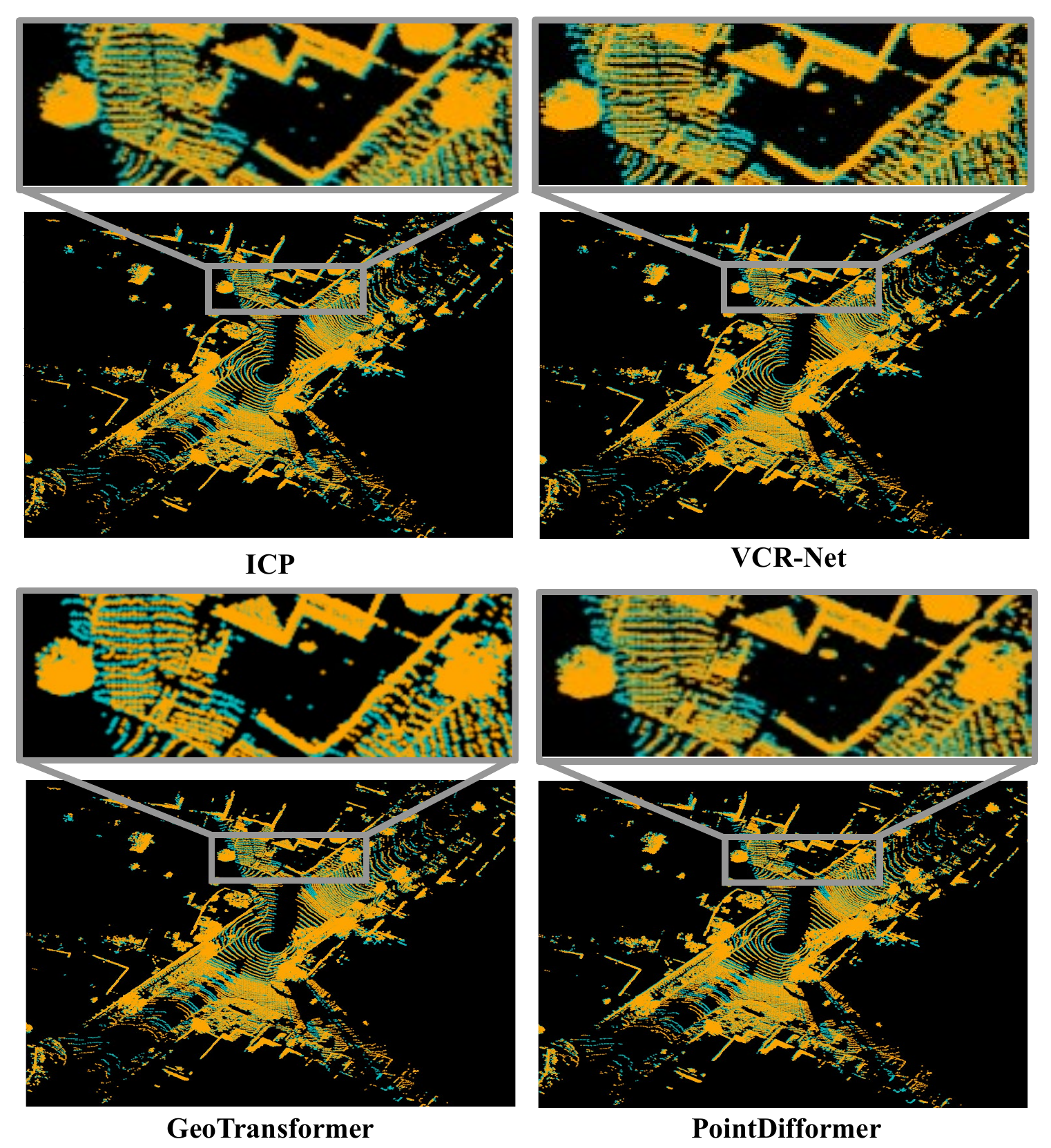}
\caption{Examples for the point cloud frame pairs from the Boreas dataset under a raining environment using different transformation prediction methods for alignment.}
\label{fig:Boreas_T_example_rain}
\end{figure}

We conduct experiments on the Boreas dataset under rainy conditions, which is considered natural noise on point clouds. From \cref{tab:Boreas_Rt_rain}, we observe that PointDifformer outperforms the baseline methods in terms of relative translation and rotation RMSEs, while having comparable performance in terms of MAEs. This indicates that our method produces fewer outliers in the predicted results, demonstrating its superior robustness compared to the baselines. We also provide several examples of point cloud alignment in \cref{fig:Boreas_T_example_rain}.

\begin{table}[!htp]
\caption{Point cloud registration performance on the Boreas dataset under the raining environment.}
\label{tab:Boreas_Rt_rain}
\centering
\newcommand{\tabincell}[2]{\begin{tabular}{@{}#1@{}}#2\end{tabular}}
\begin{tabular}{c|c c c c c} 
\hline\hline
\multirow{2}{*}{Method}         & \multicolumn{2}{c}{\tabincell{c}{Relative Translation \\ Error (centimeter [cm])}}  &  \multicolumn{2}{c}{\tabincell{c}{Relative Rotation \\ Error (degree [$^\circ$])}} & \multirow{2}{*}{\tabincell{c}{RR \\ (\%) }} \\
                                &  \textit{MAE}     & \textit{RMSE}  & \textit{MAE} & \textit{RMSE}           & \\
\hline
ICP               & 11.90               & 20.57                   & 0.15                & 0.27                & 70.8 \\ \hline
DCP               & 10.60               & 16.00                   & 0.14                & 0.22                & 75.7 \\ \hline
HGNN++            & 15.02               & 25.63                   & 0.18                & 0.32                & 52.5 \\ \hline
VCR-Net           & 8.81                & \underline{14.09}       & \underline{0.13}    & \underline{0.20}    & 84.1 \\ \hline
PCT++             & 10.39               & 16.86                   & 0.14                & 0.24                & 77.4 \\ \hline
GeoTransformer    & \textbf{4.96}       & 16.75                   & \textbf{0.10}       & 0.25                & \underline{94.9} \\ \hline
PointDifformer    & \underline{5.91}    & \textbf{8.45}           & \textbf{0.10}       & \textbf{0.14}       & \textbf{96.9} \\
\hline\hline
\end{tabular}
\end{table}

\subsection{Evaluation on Datasets with Lower Overlaps}\label{sect:3dmatch-3dlomatch}

We evaluate the performance of our method on datasets with lower overlaps, namely, 3DMatch and 3DLoMatch, as shown in \cref{tab:3DMatch_3DLoMatch}. The standard benchmark metric RR is used, and the experimental configuration follows that of \cite{qin2022geometric,yu2023rotation}.
The baselines include FCGF \cite{choy2019fully}, D3Feat \cite{bai2020d3feat}, SpinNet \cite{ao2021spinnet}, Predator \cite{huang2021predator}, YOHO \cite{wang2022you}, CoFiNet \cite{yu2021cofinet}, GeoTransformer \cite{qin2022geometric}, RoITr \cite{yu2023rotation}, and RoReg \cite{wang2023roreg}.
To perform the registration task on the 3DMatch and 3DLoMatch datasets, similar to several current state-of-the-art models like CoFiNet, GeoTransformer, RoITr and RoReg, we first employ a module to identify the areas of higher overlap in the point clouds. Specifically, we use superpoint matching \cite{qin2022geometric}. Subsequently, keypoint matching under the PointDifformer model is employed to achieve more precise correspondences for pairs of point clouds.
\Cref{tab:3DMatch_3DLoMatch} demonstrates that PointDifformer achieves state-of-the-art performance, affirming its feasibility on datasets with lower overlaps.

\begin{table}[!hbt] \footnotesize
\caption{
Performance on the 3DMatch and 3DLoMatch datasets, using the same experimental configuration as that in \cite{qin2022geometric,yu2023rotation}. 
The results of baselines are borrowed from \cite{choy2019fully,bai2020d3feat,ao2021spinnet,huang2021predator,yu2021cofinet,wang2022you,qin2022geometric,yu2023rotation,wang2023roreg}. 
}
\label{tab:3DMatch_3DLoMatch}
\centering
\newcommand{\tabincell}[2]{\begin{tabular}{@{}#1@{}}#2\end{tabular}}
\begin{tabular}{c | c  c }
\hline\hline
\textbf{Method}    & \tabincell{c}{3DMatch \\ RR (\%) \\}  & \tabincell{c}{3DLoMatch \\ RR (\%) \\}   \\ \hline
FCGF               & 85.1                      & 40.1                                           \\ \hline
D3Feat             & 81.6                      & 37.2                                           \\ \hline
SpinNet            & 88.6                      & 59.8                                           \\ \hline
Predator           & 89.0                      & 59.8                                           \\ \hline
YOHO               & 90.8                      & 65.2                                           \\ \hline
CoFiNet            & 89.3                      & 67.5                                           \\ \hline
GeoTransformer     & 92.0                      & \underline{75.0}                               \\ \hline
RoITr              & 91.9                      & 74.8                                           \\ \hline
RoReg              & \underline{92.9}          & 70.3                                           \\ \hline  
PointDifformer     & \textbf{93.0}             & \textbf{75.2}  \\    
\hline\hline 
\end{tabular}
\end{table}

\subsection{Computational Complexity}
In \cref{tab:computational_complexity}, we present the average inference time and graphics processing unit (GPU) memory required for registering each point cloud pair based on the KITTI dataset. 
We test the methods on an NVIDIA RTX A5000 GPU. The average inference time and GPU memory are measured in seconds (s) and gigabytes (GB), respectively.
From \cref{tab:computational_complexity}, we observe that PointDifformer requires higher GPU memory and incurs longer inference time compared to other baselines due to its higher complexity. On average, the inference time is still acceptable for real-time applications. A possible future work is to optimize and prune \cite{liu2018rethinking,akiba2019optuna,fang2023structural} the PointDifformer model to reduce its memory and inference time footprints.

\begin{table}[!htp]
\caption{The average inference time and GPU memory for each point cloud pair on the KITTI dataset.}
\label{tab:computational_complexity}
\centering
\resizebox{0.48\textwidth}{!}{\setlength{\tabcolsep}{4pt} 
\newcommand{\tabincell}[2]{\begin{tabular}{@{}#1@{}}#2\end{tabular}}
\begin{tabular}{c | c c c c }
\hline\hline
Method  & VCR-Net & PCT++ & GeoTransformer & PointDifformer\\ 
\hline 
\tabincell{c}{Inference time\\}
 & 0.047s & 0.648s & 0.061s & 0.072s \\ \hline 
\tabincell{c}{GPU memory\\}
& 2.29GB & 2.38GB & 1.51GB & 2.44GB \\
\hline\hline 
\end{tabular}}
\end{table}

\subsection{Ablation Study}

We perform an ablation study using the KITTI dataset under the same experimental settings as described in \cref{sect:eval_KITTI} for \cref{tab:KITTI_Rt_clear}. 
We first evaluate the efficiency of the self-attention module with heat kernel signature by comparing it with the vanilla self-attention module and the module without self-attention. As shown in \cref{tab:KITTI_Rt_ablation_hks}, the introduction of the heat kernel signature as weights into the self-attention module improves the transformation prediction accuracy. Furthermore, we observe that the vanilla self-attention module also contributes to the point cloud registration performance. 
\begin{table}[!htp]
\caption{Ablation study for the self-attention module with heat kernel signature.}
\label{tab:KITTI_Rt_ablation_hks}
\centering
\resizebox{0.48\textwidth}{!}{\setlength{\tabcolsep}{1pt} 
\newcommand{\tabincell}[2]{\begin{tabular}{@{}#1@{}}#2\end{tabular}}
\begin{tabular}{c|c c c c c} 
\hline\hline
\multirow{2}{*}{Module}         & \multicolumn{2}{c}{\tabincell{c}{Relative Translation \\ Error (centimeter [cm])}}  &  \multicolumn{2}{c}{\tabincell{c}{Relative Rotation \\ Error (degree [$^\circ$])}} &  \multirow{2}{*}{\tabincell{c}{RR \\ (\%) }} \\
                                &  \textit{MAE}     & \textit{RMSE}  & \textit{MAE} & \textit{RMSE}   & \\
\hline
Self Attention (No)                                  & 5.75               & 11.73                   & 0.16              & 0.25   
&  92.1          \\ \hline
Self Attention (Vanilla)                             & 4.56               & 10.03                   & 0.15              & 0.26   
&  94.5          \\ \hline
{\tabincell{c}{Self Attention (Heat Kernel) \\}}     & 4.14               & 8.86                    & 0.14              & 0.23   
&  97.7         \\ 
\hline\hline
\end{tabular}}
\end{table}

We investigate the influence of our proposed Point-Diffusion Net module on point cloud representation by comparing it with other graph learning methods. The results in \cref{tab:KITTI_Rt_ablation_diffusion} show that the point cloud registration model with our Point-Diffusion Net outperforms those with other graph learning methods including HGNN \cite{feng2019hypergraph} and DGCNN \cite{wang2019dynamic}. This indicates that the Point-Diffusion Net can achieve a more robust representation of the point cloud. 
\begin{table}[!htp]
\caption{Ablation study for the effectiveness of Point-Diffusion Net.}
\label{tab:KITTI_Rt_ablation_diffusion}
\centering
\resizebox{0.48\textwidth}{!}{\setlength{\tabcolsep}{2pt} 
\newcommand{\tabincell}[2]{\begin{tabular}{@{}#1@{}}#2\end{tabular}}
\begin{tabular}{c|c c c c c} 
\hline\hline
\multirow{2}{*}{\tabincell{c}{Method}}         & \multicolumn{2}{c}{\tabincell{c}{Relative Translation \\ Error (centimeter [cm])}}  &  \multicolumn{2}{c}{\tabincell{c}{Relative Rotation \\ Error (degree [$^\circ$])}} &  \multirow{2}{*}{\tabincell{c}{RR \\ (\%) }} \\
                                &  \textit{MAE}     & \textit{RMSE}  & \textit{MAE} & \textit{RMSE}  & \\
\hline
HGNN                    & 8.22               & 16.40                  & 0.19                & 0.31   &  90.1            \\ \hline
DGCNN                   & 4.46               & 9.55                   & 0.14                & 0.23   &  97.3            \\ \hline
Point-Diffusion Net     & 4.14               & 8.86                   & 0.14                & 0.23   &  97.7            \\ 
\hline\hline
\end{tabular}}
\end{table}

We investigate the impact of the number of selected top $K'$ corresponding keypoints from the attention-based correspondence module. Specifically, we select $25\%$, $50\%$, $75\%$, and $100\%$ corresponding keypoints from the entire set of keypoints. The results in \cref{tab:KITTI_Rt_ablation_keypoint_K'} show that selecting $50\%$ and $75\%$ corresponding keypoints achieves better performance.

\begin{table}[!htp]
\caption{Ablation study for the number of selected corresponding keypoints in the attention-based correspondence.}
\label{tab:KITTI_Rt_ablation_keypoint_K'}
\centering
\newcommand{\tabincell}[2]{\begin{tabular}{@{}#1@{}}#2\end{tabular}}
\begin{tabular}{c|c c c c c c} 
\hline\hline
\multirow{2}{*}{\tabincell{c}{Proportion}}         & \multicolumn{2}{c}{\tabincell{c}{Relative Translation \\ Error (centimeter [cm])}}  &  \multicolumn{2}{c}{\tabincell{c}{Relative Rotation \\ Error (degree [$^\circ$])}} &  \multirow{2}{*}{\tabincell{c}{RR \\ (\%) }} \\
                                &  \textit{MAE}     & \textit{RMSE}  & \textit{MAE} & \textit{RMSE}  & \\
\hline
$25\%$             & 8.88               & 15.84                  & 0.20                & 0.32     & 88.1           \\ \hline
$50\%$             & 4.06               & 7.95                   & 0.15                & 0.25     & 97.3           \\ \hline
$75\%$             & 4.14               & 8.86                   & 0.14                & 0.23     & 97.7           \\ \hline
$100\%$            & 7.20               & 16.14                  & 0.19                & 0.34     & 92.2           \\ 
\hline\hline
\end{tabular}
\end{table}

To evaluate the effectiveness of our total loss $\calL_{\mathrm{total}}$ in \cref{eq:Loss_total}, we compare it with the corresponding point loss $\calL_{\mathrm{point}}$ in \cref{eq:Loss_p2p} and the ground-truth-to-prediction loss $\calL_{\mathrm{rt}}$ in \cref{eq:Loss_rt}. Based on the results presented in \cref{tab:KITTI_Rt_ablation_loss}, we observe that the total loss $\calL_{\mathrm{total}}$ outperforms the others, suggesting that incorporating more information in the loss function has a positive effect on training. Additionally, we find that $\calL_{\mathrm{point}}$ surpasses $\calL_{\mathrm{rt}}$, which implies that the point loss plays a more important role in the $\calL_{\mathrm{total}}$. 
\begin{table}[!htp]
\caption{Ablation study for the loss function.}
\label{tab:KITTI_Rt_ablation_loss}
\centering
\newcommand{\tabincell}[2]{\begin{tabular}{@{}#1@{}}#2\end{tabular}}
\begin{tabular}{c|c c c c c} 
\hline\hline
\multirow{2}{*}{Loss}         & \multicolumn{2}{c}{\tabincell{c}{Relative Translation \\ Error (centimeter [cm])}}  &  \multicolumn{2}{c}{\tabincell{c}{Relative Rotation \\ Error (degree [$^\circ$])}} &  \multirow{2}{*}{\tabincell{c}{RR \\ (\%) }}\\
                                &  \textit{MAE}     & \textit{RMSE}  & \textit{MAE} & \textit{RMSE}  & \\
\hline
$\calL_{\mathrm{rt}}$         & 6.32               & 11.55                  & 0.21               & 0.39    &  94.2    \\ \hline
$\calL_{\mathrm{point}}$      & 5.41               & 11.44                  & 0.15               & 0.24    &  96.3    \\ \hline
$\calL_{\mathrm{total}}$      & 4.14               & 8.86                   & 0.14               & 0.23    &  97.7    \\ 
\hline\hline
\end{tabular}
\end{table}

\section{Conclusions}\label{sect:conc}

In order to develop a robust 3D point cloud registration approach that is able to handle noise or perturbations, we have utilized graph neural PDE modules to learn point cloud feature representations. We have also designed attention modules with heat kernel signatures to establish correspondence between points from two point clouds. Our approach has been extensively evaluated through experiments, which demonstrate that it generally outperforms baselines not only on raw point clouds but also on point clouds with additive noise and 3D shape perturbations. These results suggest that graph neural PDEs are beneficial for the task of point cloud registration.

In this paper, we have conducted a robustness study limited only to perturbations through Gaussian noise, rain, and partial removal of a frame. A future work of interest is to further investigate the robustness of our method under diverse perturbations, including \emph{adversarial attacks}. Furthermore, we aim to enhance our model's adaptability to noisy datasets under different environmental factors.

\section*{ACKNOWLEDGMENT}
To improve the readability, parts of this paper have been grammatically revised using ChatGPT \cite{OpenAI}.

%
%
%

\bibliographystyle{IEEEtran}
\bibliography{IEEEabrv,StringDefinitions,refs}

\vfill

\end{document}

%% file: preamble.tex
\usepackage[T1]{fontenc}
\usepackage[utf8]{inputenc}
\usepackage{mathtools}

\usepackage{amssymb,mathrsfs}
\usepackage{amsthm}
\usepackage{bm}
\usepackage{scalerel}
\usepackage{nicefrac}
\usepackage{microtype} 
\usepackage[shortlabels]{enumitem}
\usepackage{graphicx}
\usepackage{epstopdf}
\DeclareGraphicsExtensions{.eps,.png,.jpg,.pdf}

\usepackage{url}
\usepackage{colortbl}
\usepackage{booktabs}
\usepackage{multirow}
\usepackage{colortbl,xcolor}
\usepackage[normalem]{ulem}
\usepackage{xparse,xstring}
\usepackage{calc}
\usepackage{etoolbox}

\makeatletter
\@ifpackageloaded{natbib}{
	\relax
}{
	\usepackage{cite}
}
\makeatother

\usepackage{array}
\newcolumntype{L}[1]{>{\raggedright\let\newline\\\arraybackslash\hspace{0pt}}m{#1}}
\newcolumntype{C}[1]{>{\centering\let\newline\\\arraybackslash\hspace{0pt}}m{#1}}
\newcolumntype{R}[1]{>{\raggedleft\let\newline\\\arraybackslash\hspace{0pt}}m{#1}}

\makeatletter
\let\MYcaption\@makecaption
\makeatother
\usepackage[font=footnotesize]{subcaption}
\makeatletter
\let\@makecaption\MYcaption
\makeatother

\usepackage{glossaries}
\makeatletter
\sfcode`\.1006

\let\oldgls\gls
\let\oldglspl\glspl

\newcommand\fussy@ifnextchar[3]{%
	\let\reserved@d=#1%
	\def\reserved@a{#2}%
	\def\reserved@b{#3}%
	\futurelet\@let@token\fussy@ifnch}
\def\fussy@ifnch{%
	\ifx\@let@token\reserved@d
		\let\reserved@c\reserved@a
	\else
		\let\reserved@c\reserved@b
	\fi
	\reserved@c}

\renewcommand{\gls}[1]{%
\oldgls{#1}\fussy@ifnextchar.{\@checkperiod}{\@}}
\renewcommand{\glspl}[1]{%
\oldglspl{#1}\fussy@ifnextchar.{\@checkperiod}{\@}}

\newcommand{\@checkperiod}[1]{%
	\ifnum\sfcode`\.=\spacefactor\else#1\fi
}

\robustify{\gls}
\robustify{\glspl}
\makeatother

\newacronym{wrt}{w.r.t.}{with respect to}
\newacronym{RHS}{R.H.S.}{right-hand side}
\newacronym{LHS}{L.H.S.}{left-hand side}
\newacronym{iid}{i.i.d.}{independent and identically distributed}
\newacronym{SOTA}{SOTA}{state-of-the-art}

\usepackage{float}

\ifx\notloadhyperref\undefined
	\ifx\loadbibentry\undefined
		\usepackage[hidelinks,hypertexnames=false]{hyperref}
	\else
		\usepackage{bibentry}
		\makeatletter\let\saved@bibitem\@bibitem\makeatother
		\usepackage[hidelinks,hypertexnames=false]{hyperref}
		\makeatletter\let\@bibitem\saved@bibitem\makeatother
	\fi
\else
	\ifx\loadbibentry\undefined
		\relax
	\else
		\usepackage{bibentry}
	\fi
\fi

\usepackage[capitalize]{cleveref}
\crefname{equation}{}{}
\Crefname{equation}{}{}
\crefname{claim}{claim}{claims}
\crefname{step}{step}{steps}
\crefname{line}{line}{lines}
\crefname{condition}{condition}{conditions}
\crefname{dmath}{}{}
\crefname{dseries}{}{}
\crefname{dgroup}{}{}

\crefname{Problem}{Problem}{Problems}
\crefformat{Problem}{Problem~#2#1#3}
\crefrangeformat{Problem}{Problems~#3#1#4 to~#5#2#6}

\crefname{Theorem}{Theorem}{Theorems}
\crefname{Corollary}{Corollary}{Corollaries}
\crefname{Proposition}{Proposition}{Propositions}
\crefname{Lemma}{Lemma}{Lemmas}
\crefname{Definition}{Definition}{Definitions}
\crefname{Example}{Example}{Examples}
\crefname{Assumption}{Assumption}{Assumptions}
\crefname{Remark}{Remark}{Remarks}
\crefname{Rem}{Remark}{Remarks}
\crefname{remarks}{Remarks}{Remarks}
\crefname{Appendix}{Appendix}{Appendices}
\crefname{Supplement}{Supplement}{Supplements}
\crefname{Exercise}{Exercise}{Exercises}
\crefname{Theorem_A}{Theorem}{Theorems}
\crefname{Corollary_A}{Corollary}{Corollaries}
\crefname{Proposition_A}{Proposition}{Propositions}
\crefname{Lemma_A}{Lemma}{Lemmas}
\crefname{Definition_A}{Definition}{Definitions}

\usepackage{crossreftools}
\ifx\notloadhyperref\undefined
\else
	\relax
\fi

\usepackage{algorithm}
\usepackage{algpseudocode}

\ifx\loadbreqn\undefined
	\relax
\else
	\usepackage{breqn}
\fi

\makeatletter
\def\cleartheorem#1{%
    \expandafter\let\csname#1\endcsname\relax
    \expandafter\let\csname c@#1\endcsname\relax
}
\def\clearthms#1{ \@for\tname:=#1\do{\cleartheorem\tname} }
\makeatother

\ifx\renewtheorem\undefined
	\ifx\useTheoremCounter\undefined
		\newtheorem{Theorem}{Theorem}
		\newtheorem{Corollary}{Corollary}
		\newtheorem{Proposition}{Proposition}
		
	\else

	\fi

\fi

\theoremstyle{remark}

\theoremstyle{plain}

\newcommand{\qednew}{\nobreak \ifvmode \relax \else
		\ifdim\lastskip<1.5em \hskip-\lastskip
			\hskip1.5em plus0em minus0.5em \fi \nobreak
		\vrule height0.75em width0.5em depth0.25em\fi}



\newcommand{\nn}{\nonumber\\ }

\NewDocumentCommand{\movedownsub}{e{^_}}{%
	\IfNoValueTF{#1}{%
		\IfNoValueF{#2}{^{}}
	}{%
		^{#1}
	}%
	\IfNoValueF{#2}{_{#2}}
}

\let\latexchi\chi
\RenewDocumentCommand{\chi}{}{\latexchi\movedownsub}

\newcommand{\Real}{\mathbb{R}}

\newcommand{\calL}{\mathcal{L}}

\newcommand{\calN}{\mathcal{N}}

\newcommand{\bI}{\mathbf{I}}

\newcommand{\bM}{\mathbf{M}}

\newcommand{\bR}{\mathbf{R}}

\newcommand{\bt}{\mathbf{t}}

\newcommand{\bU}{\mathbf{U}}

\newcommand{\bV}{\mathbf{V}}
\newcommand{\bw}{\mathbf{w}}
\newcommand{\bW}{\mathbf{W}}
\newcommand{\bx}{\mathbf{x}}
\newcommand{\bX}{\mathbf{X}}
\newcommand{\by}{\mathbf{y}}
\newcommand{\bY}{\mathbf{Y}}

\newcommand{\bZ}{\mathbf{Z}}

\DeclareSymbolFont{bsfletters}{OT1}{cmss}{bx}{n}
\DeclareSymbolFont{ssfletters}{OT1}{cmss}{m}{n}
\DeclareMathSymbol{\bsfGamma}{0}{bsfletters}{'000}
\DeclareMathSymbol{\ssfGamma}{0}{ssfletters}{'000}
\DeclareMathSymbol{\bsfDelta}{0}{bsfletters}{'001}
\DeclareMathSymbol{\ssfDelta}{0}{ssfletters}{'001}
\DeclareMathSymbol{\bsfTheta}{0}{bsfletters}{'002}
\DeclareMathSymbol{\ssfTheta}{0}{ssfletters}{'002}
\DeclareMathSymbol{\bsfLambda}{0}{bsfletters}{'003}
\DeclareMathSymbol{\ssfLambda}{0}{ssfletters}{'003}
\DeclareMathSymbol{\bsfXi}{0}{bsfletters}{'004}
\DeclareMathSymbol{\ssfXi}{0}{ssfletters}{'004}
\DeclareMathSymbol{\bsfPi}{0}{bsfletters}{'005}
\DeclareMathSymbol{\ssfPi}{0}{ssfletters}{'005}
\DeclareMathSymbol{\bsfSigma}{0}{bsfletters}{'006}
\DeclareMathSymbol{\ssfSigma}{0}{ssfletters}{'006}
\DeclareMathSymbol{\bsfUpsilon}{0}{bsfletters}{'007}
\DeclareMathSymbol{\ssfUpsilon}{0}{ssfletters}{'007}
\DeclareMathSymbol{\bsfPhi}{0}{bsfletters}{'010}
\DeclareMathSymbol{\ssfPhi}{0}{ssfletters}{'010}
\DeclareMathSymbol{\bsfPsi}{0}{bsfletters}{'011}
\DeclareMathSymbol{\ssfPsi}{0}{ssfletters}{'011}
\DeclareMathSymbol{\bsfOmega}{0}{bsfletters}{'012}
\DeclareMathSymbol{\ssfOmega}{0}{ssfletters}{'012}

\makeatletter
\newcommand*\rel@kern[1]{\kern#1\dimexpr\macc@kerna}
\newcommand*\widebar[1]{%
  \begingroup
  \def\mathaccent##1##2{%
    \rel@kern{0.8}%
    \overline{\rel@kern{-0.8}\macc@nucleus\rel@kern{0.2}}%
    \rel@kern{-0.2}%
  }%
  \macc@depth\@ne
  \let\math@bgroup\@empty \let\math@egroup\macc@set@skewchar
  \mathsurround\z@ \frozen@everymath{\mathgroup\macc@group\relax}%
  \macc@set@skewchar\relax
  \let\mathaccentV\macc@nested@a
  \macc@nested@a\relax111{#1}%
  \endgroup
}
\makeatother

\DeclareMathOperator*{\argmin}{arg\,min}

\DeclareMathOperator{\var}{var}

\DeclareMathOperator{\cov}{cov}

\DeclareMathOperator*{\concat}{\scalerel*{\parallel}{\sum}}

\newcommand{\ifbcdot}[1]{\ifblank{#1}{\cdot}{#1}}

\DeclarePairedDelimiterX\abs[1]{\lvert}{\rvert}{\ifbcdot{#1}}
\DeclarePairedDelimiterX\parens[1]{(}{)}{\ifbcdot{#1}}
\DeclarePairedDelimiterX\brk[1]{[}{]}{\ifbcdot{#1}}
\DeclarePairedDelimiterX\braces[1]{\{}{\}}{\ifbcdot{#1}}
\DeclarePairedDelimiterX\angles[1]{\langle}{\rangle}{\ifblank{#1}{\cdot,\cdot}{#1}}
\DeclarePairedDelimiterX\ip[2]{\langle}{\rangle}{\ifbcdot{#1},\ifbcdot{#2}}
\DeclarePairedDelimiterX\norm[1]{\lVert}{\rVert}{\ifbcdot{#1}}
\DeclarePairedDelimiterX\ceil[1]{\lceil}{\rceil}{\ifbcdot{#1}}
\DeclarePairedDelimiterX\floor[1]{\lfloor}{\rfloor}{\ifbcdot{#1}}

\DeclareFontFamily{U}{matha}{\hyphenchar\font45}
\DeclareFontShape{U}{matha}{m}{n}{
      <5> <6> <7> <8> <9> <10> gen * matha
      <10.95> matha10 <12> <14.4> <17.28> <20.74> <24.88> matha12
      }{}
\DeclareSymbolFont{matha}{U}{matha}{m}{n}
\DeclareFontSubstitution{U}{matha}{m}{n}

\DeclareFontFamily{U}{mathx}{\hyphenchar\font45}
\DeclareFontShape{U}{mathx}{m}{n}{
      <5> <6> <7> <8> <9> <10>
      <10.95> <12> <14.4> <17.28> <20.74> <24.88>
      mathx10
      }{}
\DeclareSymbolFont{mathx}{U}{mathx}{m}{n}
\DeclareFontSubstitution{U}{mathx}{m}{n}

\DeclareMathDelimiter{\vvvert}{0}{matha}{"7E}{mathx}{"17}
\DeclarePairedDelimiterX\vertiii[1]{\vvvert}{\vvvert}{\ifbcdot{#1}}

\DeclarePairedDelimiterXPP\trace[1]{\operatorname{Tr}}{(}{)}{}{\ifbcdot{#1}} 
\DeclarePairedDelimiterXPP\col[1]{\operatorname{col}}{\{}{\}}{}{\ifbcdot{#1}} 
\DeclarePairedDelimiterXPP\row[1]{\operatorname{row}}{\{}{\}}{}{\ifbcdot{#1}} 
\DeclarePairedDelimiterXPP\erf[1]{\operatorname{erf}}{(}{)}{}{\ifbcdot{#1}}
\DeclarePairedDelimiterXPP\erfc[1]{\operatorname{erfc}}{(}{)}{}{\ifbcdot{#1}}
\DeclarePairedDelimiterXPP\KLD[2]{D}{(}{)}{}{\ifbcdot{#1}\, \delimsize\|\, \ifbcdot{#2}} 
\DeclarePairedDelimiterXPP\op[2]{\operatorname{#1}}{(}{)}{}{#2} 

\newcommand{\T}{^{\mkern-1.5mu\mathop\intercal}}

\DeclarePairedDelimiterXPP\indicate[1]{{\bf 1}}{\{}{\}}{}{\ifbcdot{#1}}

\NewDocumentCommand\ofrac{s m}{%
	\IfBooleanTF#1%
	{\dfrac{1}{#2}}%
	{\frac{1}{#2}}%
}
\NewDocumentCommand\ddfrac{s m m}{%
	\IfBooleanTF#1%
	{\dfrac{\mathrm{d} {#2}}{\mathrm{d} {#3}}}%
	{\frac{\mathrm{d} {#2}}{\mathrm{d} {#3}}}%
}
\NewDocumentCommand\ppfrac{s m m}{%
	\IfBooleanTF#1%
	{\dfrac{\partial {#2}}{\partial {#3}}}%
	{\frac{\partial {#2}}{\partial {#3}}}%
}

\providecommand\given{}

\DeclarePairedDelimiterX\Set[2]\{\}{%
\renewcommand\given{\SetSymbol[\delimsize]{#1}}
#2
}
\DeclarePairedDelimiterX\Setc[1]\{\}{%
\renewcommand\given{\SetSymbol{:}}
#1
}

\NewDocumentCommand\set{s o m}{%
	\IfBooleanTF#1%
	{\IfValueTF{#2}{\Set*{#2}{#3}}{\Setc*{#3}}}%
	{\IfValueTF{#2}{\Set{#2}{#3}}{\Setc{#3}}}%
}

\NewDocumentCommand{\evalat}{ s O{\big} m e{_^} }{%
\IfBooleanTF{#1}%
{\left. #3 \right|}{#3#2|}%
\IfValueT{#4}{_{#4}}%
\IfValueT{#5}{^{#5}}%
}

\providecommand\given{}
\DeclarePairedDelimiterXPP\cprob[1]{}(){}{
\renewcommand\given{\nonscript\,\delimsize\vert\allowbreak\nonscript\,\mathopen{}}%
#1%
}
\DeclarePairedDelimiterXPP\cexp[1]{}[]{}{
\renewcommand\given{\nonscript\,\delimsize\vert\allowbreak\nonscript\,\mathopen{}}%
#1%
}

\DeclareDocumentCommand \P { s e{_^} d() g } {%
	\mathbb{P}%
	\IfBooleanTF{#1}%
		{
			\IfValueT{#2}{_{#2}}%
			\IfValueT{#3}{^{#3}}%
			\IfValueTF{#5}{\cprob{#4 \given #5}}{\IfValueT{#4}{\cprob{#4}}}%
		}%
		{
			\IfValueT{#2}{_{#2}}%
			\IfValueT{#3}{^{#3}}%
			\IfValueTF{#5}{\cprob*{#4 \given #5}}{\IfValueT{#4}{\cprob*{#4}}}%
		}%
}

\DeclareDocumentCommand \E { s e{_^} o g } {%
	\mathbb{E}%
	\IfBooleanTF{#1}%
		{
			\IfValueT{#2}{_{#2}}%
			\IfValueT{#3}{^{#3}}%
			\IfValueTF{#5}{\cexp{#4 \given #5}}{\IfValueT{#4}{\cexp{#4}}}%
		}%
		{
			\IfValueT{#2}{_{#2}}%
			\IfValueT{#3}{^{#3}}%
			\IfValueTF{#5}{\cexp*{#4 \given #5}}{\IfValueT{#4}{\cexp*{#4}}}%
		}%
}

\DeclareDocumentCommand \Var { s e{_^} d() g } {%
	\var%
	\IfBooleanTF{#1}%
		{
			\IfValueT{#2}{_{#2}}%
			\IfValueT{#3}{^{#3}}%
			\IfValueTF{#5}{\cprob{#4 \given #5}}{\IfValueT{#4}{\cprob{#4}}}%
		}%
		{
			\IfValueT{#2}{_{#2}}%
			\IfValueT{#3}{^{#3}}%
			\IfValueTF{#5}{\cprob*{#4 \given #5}}{\IfValueT{#4}{\cprob*{#4}}}%
		}%
}

\DeclareDocumentCommand \Cov { s e{_^} d() g } {%
	\cov%
	\IfBooleanTF{#1}%
		{
			\IfValueT{#2}{_{#2}}%
			\IfValueT{#3}{^{#3}}%
			\IfValueTF{#5}{\cprob{#4 \given #5}}{\IfValueT{#4}{\cprob{#4}}}%
		}%
		{
			\IfValueT{#2}{_{#2}}%
			\IfValueT{#3}{^{#3}}%
			\IfValueTF{#5}{\cprob*{#4 \given #5}}{\IfValueT{#4}{\cprob*{#4}}}%
		}%
}

\ExplSyntaxOn
\NewDocumentCommand \dist {m o o} {%
\mathrm{#1}\left(%
	\IfValueT{#3}{%
		\tl_if_blank:nTF{ #3 }{\cdot\, \middle|\, }{#3\, \middle|\, }%
	}
	\IfValueT{#2}{#2}%
\right)%
}
\ExplSyntaxOff

\NewDocumentCommand {\cbrace} {t+ D[]{black} D(){\widthof{#5}} m m } {%
	\begingroup%
		\color{#2}
		\IfBooleanTF{#1}{%
			\overbrace{#4}^%
		}{
			\underbrace{#4}_%
		}%
		{\parbox[c]{#3}{\centering\footnotesize{#5}}}%
	\endgroup%
}

\let\oldforall\forall
\renewcommand{\forall}{\oldforall \, }

\let\oldexist\exists
\renewcommand{\exists}{\oldexist \, }

\makeatletter

\newcommand{\rankcolor}[2]{%
	\expandafter\renewcommand\csname #1\endcsname[1]{%
		\ifblank{##1}{%
			{\color{#2} \textbf{#2}}%
		}{%
			\ifmmode
				\textcolor{#2}{\bm{##1}}%
			\else%
				{\color{#2} \textbf{##1}}%
			\fi	
		}%
	}
}

\rankcolor{first}{red}
\rankcolor{second}{blue}
\rankcolor{third}{cyan}
\makeatother

\newcommand{\figref}[1]{Fig.~\ref{#1}}
\graphicspath{{./Figures/}{./figures/}}
\DeclareDocumentCommand{\includeCroppedPdf}{ o O{./Figures/} m }{
	\IfFileExists{#2#3-crop.pdf}{}{%
		\immediate\write18{pdfcrop #2#3.pdf #2#3-crop.pdf}}%
	\includegraphics[#1]{#2#3-crop.pdf}
}

\makeatletter
\newcommand*{\addFileDependency}[1]{
  \typeout{(#1)}
  \@addtofilelist{#1}
  \IfFileExists{#1}{}{\typeout{No file #1.}}
}
\makeatother

\definecolor{gray90}{gray}{0.9}
\def\colorlist{red,blue,brown,cyan,darkgray,gray,lightgray,green,lime,magenta,olive,orange,pink,purple,teal,violet,white,yellow}

\makeatletter
\def\startcomment{[}
\ifx\nohighlights\undefined
	\newcommand{\createcolor}[1]{%
			\expandafter\newcommand\csname #1\endcsname[1]{{\color{#1} ##1}}%
	}
	\newcommand{\msout}[1]{\text{\color{green} \sout{\ensuremath{#1}}}}
	\newcommand{\del}[1]{{\color{green}\ifmmode \msout{#1}\else\sout{#1}\fi}}
\else
	\newcommand{\createcolor}[1]{%
			\expandafter\newcommand\csname #1\endcsname[1]{%
				\noexpandarg%
				\StrChar{##1}{1}[\firstletter]%
				\if\firstletter\startcomment%
					\relax
				\else%
					##1
				\fi
			}%
	}
	\newcommand{\msout}[1]{}
	\newcommand{\del}[1]{}
\fi

\def\@tempa#1,{%
    \ifx\relax#1\relax\else
        \createcolor{#1}%
        \expandafter\@tempa
    \fi
}
\expandafter\@tempa\colorlist,\relax,
\makeatother

\newcommand{\hhide}[1]{}

\ifx\diagnoselabel\undefined
	\relax
\else
	\makeatletter
	\def\@testdef #1#2#3{%
		\def\reserved@a{#3}\expandafter \ifx \csname #1@#2\endcsname
			\reserved@a  \else
			\typeout{^^Jlabel #2 changed:^^J%
				\meaning\reserved@a^^J%
				\expandafter\meaning\csname #1@#2\endcsname^^J}%
			\@tempswatrue \fi}
	\makeatother
\fi
